\documentclass[twoside,10pt]{article}

\title{Logic-Based Artificial Intelligence Algorithms\\ Supporting Categorical Semantics}

\usepackage{fancyhdr}
\usepackage{latexsym}
\usepackage{pifont}       
\usepackage{amsmath}
\usepackage{amsfonts}

\usepackage{graphicx}     
\usepackage[T1]{fontenc}
\usepackage{ifthen}
\usepackage{pstricks,pst-grad,pst-text,pst-node,multido,pst-plot,calc,pst-3dplot}

\newcommand{\sequent}[3]{\left(\ensuremath{{#1}\;\vdash_{{#2}}{#3}\right)}}
\newcommand{\sequentS}[3]{\left(\ensuremath{{#1}\;\dashv\,\vdash_{{#2}}{#3}\right)}}
\newcommand{\myproof}{\noindent\textit{Because}: }
\newcommand{\interp}[1]{\ensuremath[\![{#1}]\!]}
\newcommand{\Set}{\mbox{Set}}
\newcommand{\interpretation}[1]{\ensuremath{[\![{#1}]\!]}}
\usepackage{amssymb}
\usepackage{amsmath}
\usepackage{algorithm}
\usepackage[noend]{algpseudocode}
\usepackage[all]{xy}
\usepackage{graphicx}
\usepackage{pstricks,pst-grad,pst-text,pst-node,multido,pst-plot,calc,pst-3dplot,pst-func}
\usepackage{wrapfig}

\newcommand{\function}[1]{\textsc{\color{black}{#1}}}

\definecolor{MyLinkColor}{rgb}{.1,.1,.7}
\definecolor{MyCiteColor}{rgb}{.1,.7,.1}
\definecolor{MyURLColor}{rgb}{.2,.2,.7}
\usepackage[backref=true,pagebackref=false,hyperindex,colorlinks=true,
  linkcolor=MyLinkColor,citecolor=MyCiteColor,urlcolor=MyURLColor]{hyperref}

\usepackage{amsthm}
\newtheorem{theorem}{Theorem}[section]
\newtheorem{lemma}[theorem]{Lemma}
\newcommand{\enpr}{$\square$}

\definecolor{lgray}{rgb}{0.8,0.8,0.8}
\definecolor{edit}{rgb}{0,0,0} 

\begin{document}
\pagestyle{empty}
{\ }
\setlength{\baselineskip}{1.07\baselineskip}

\begin{center}
\textbf{\LARGE Logic-Based Artificial Intelligence Algorithms}
\smallskip

\textbf{\LARGE Supporting Categorical Semantics}
\vspace{.3in}

\begin{tabular}{c}
\large Ralph L. Wojtowicz\\[3pt]
\small Shenandoah University\\[-2pt]
\small Departmenet of Mathematics\\[-2pt]
\small Winchester, VA  USA\\[2pt]
rwojtowi@su.edu, ralphw@bakermountain.org
\end{tabular}
\end{center}
\vspace{.1in}

\begin{abstract}
This paper seeks to apply categorical logic to the design of artificial intelligent
agents that reason symbolically about objects more richly structured than sets.
Using Johnstone's sequent calculus of terms- and formulae-in-context, we develop
forward chaining and normal form  algorithms 
for reasoning about objects in cartesian categories with the rules for Horn logic.
We also adapt first-order unification 
 to support multi-sorted theories, contexts, and 
fragments of first-order logic.
The significance of these reformulations rests in the fact that they can be applied to 
reasoning about objects in semantic categories that do not support classical logic
or even all its connectives.


\end{abstract}

\section{Introduction}\label{sec:introduction}

The discovery of categorical logic is one of the  20${^{\mbox{\scriptsize th}}}$
century's great contributions to mathematics.
A facet of 
this field is characterizations of the semantic categories for 
classes of logical theories.
First-order Horn, regular, coherent and intuitionistic theories, for example,
are respectively interpreted in  cartesian, regular, coherent, and Heyting 
  categories~\cite{Bell, CA, Elephant}.
Topological spaces and sheaves provide semantics for propositional and 
first-order S4 modal logics~\cite{AwKo}.
Cartesian-closed categories give semantics for the typed $\lambda$-calculus~\cite{Elephant, Lambek}.
Fragments of linear logic are interpreted in  $*$-autonomous categories~\cite{Barr}. 

This paper grew from an effort to use the syntax and categorical semantics of~\cite{Elephant}
to reformulate the logic-based artificial intelligence (AI) methods of~\cite{RN}.
It is motivated by both pedagogy and applications.
Over the past two years, we have 
taught undergraduate AI courses using this formulation and have found
the sequent calculus of~\cite{Elephant} (see~[\ref{sec:deduction}] of this paper)
to have several advantages. 
It is concise, has precise rules for substitution and equality,
and can be introduced incrementally using  fragments of first-order logic.
Derived axioms serve as exercises.  
Categorical logic also clarifies the distinction between syntax and semantics.  
Truth tables and Venn diagrams, for example, arise as 
propositional semantics in $\Set$ and $\Set/X$.  
Directed graphs and other familiar objects 
illustrate semantics with multiple truth values.

\begin{wrapfigure}[7]{R}{5.9cm}
\begin{center}
\begin{pspicture}(0.5,0.0)(5,2.15)
\psset{arrowsize=3pt 1, nodesep=2pt, linewidth=0.02}
\psframe[fillstyle=solid,fillcolor=lgray,framearc=0.3](-0.1,0)(3.1,3.0)
\rput{90}(0.25,1.5){Agent}
\rput[b](1.5,2.6){\small Sensors}
\rput[b](1.5,0.2){\small Actuators}  
\psframe[fillstyle=solid,fillcolor=white,framearc=0.3](0.75,0.75)(2.25,2.25)
  \psline{->}(1.5,2.5)(1.5,1.9)  
  \psline{->}(1.5,1.0)(1.5,0.45) 
  \rput(1.5,1.5){\small \begin{tabular}{c}Agent\\[-1pt] Program\end{tabular}}
\psframe[fillstyle=solid,fillcolor=lgray,framearc=0.3](4.6,0)(5.7,3.0)
\rput{90}(5.15,1.5){Environment}
\psline{->}(2.4,0.3)(4.95,0.3) 
\psline{<-}(2.4,2.65)(4.95,2.65) 
\rput[t](3.775,0.2){\small Actions}
\rput[b](3.775,2.75){\small Percepts}

\end{pspicture}
\end{center}
\end{wrapfigure}
\cite{RN} formulates the field of  AI in terms of agents.  
An \textit{agent} is a system that can be viewed as perceiving its  environment 
through 
sensors and acting upon the environment through
actuators.  The engineer's task is to design and implement 
an \textit{agent program} that processes inputs then
 chooses appropriate actions.  
Agents may  maintain representations of goals, history,
and  the environment.
\textit{Logic-based agents} use representations expressed
as logical theories. 
Traditional presentations  (e.g.,~\cite{Luger, Mitchell, RN})
rely on classical logic and  semantics.



This paper seeks to apply categorical logic to the 
design of  agents that reason symbolically about objects more richly structured
than sets.  
Use of such abstraction barriers is an idiom that is frequently used to manage 
 software complexity~\cite{Wizard}.
%
As an illustration,
consider a theory $\mathbb{T}$ with two sorts, Points and Lines, a function 
symbol $f:\mbox{Points},\mbox{Points}\to \mbox{Lines}$, a relation symbol
$I\rightarrowtail \mbox{Points}, \mbox{Lines}$, and  an axiom
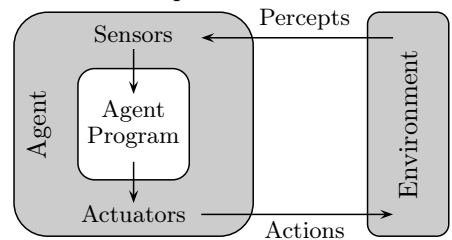
\vfill
\eject
\pagestyle{fancy}

\noindent
$\sequent{\top}{x,y}{(I(x, f(x,y))\wedge I(y, f(x,y)))}$. 
We define a model $M$ of $\mathbb{T}$
in the category $\Set^{\circlearrowleft}$  (\cite{LS, Woj04}). 
Objects of $\Set^{\circlearrowleft}$ are \textit{iterators} (discrete-time dynamic systems): pairs $(X, g)$ with $X$
a set and $g:X\to X$ a function.  
A map  $\phi:(X,\, g)\to (Y,\,h)$ 
is a function that respects 
the dynamics:  $\varphi\circ g = h\circ \varphi$.  
We interpret the  sorts of $\mathbb{T}$ as iterators $M(\mbox{Points})$ and $M(\mbox{Lines})$, the 
function symbol as a map $M(f):M(\mbox{Points})\times M(\mbox{Points})\to M(\mbox{Lines})$
and the relation symbol as a subobject $M(I)\leq M(\mbox{Points})\times M(\mbox{Lines})$.
For $x\in M(\mbox{Points})$, let $t_p$ be the time for $p$ to enter the 2-cycle.  
Points $p$ and $q$ are on line $e_i$ if $t_p-t_q$ is even and $\mbox{max}(t_p, t_q)=i$.
They are on line $o_i$ if $t_p-t_q$ is odd. 
$x$ and $y$ are on the line $o_1$, for example, while 
$w$ and $y$ are on $e_2$.

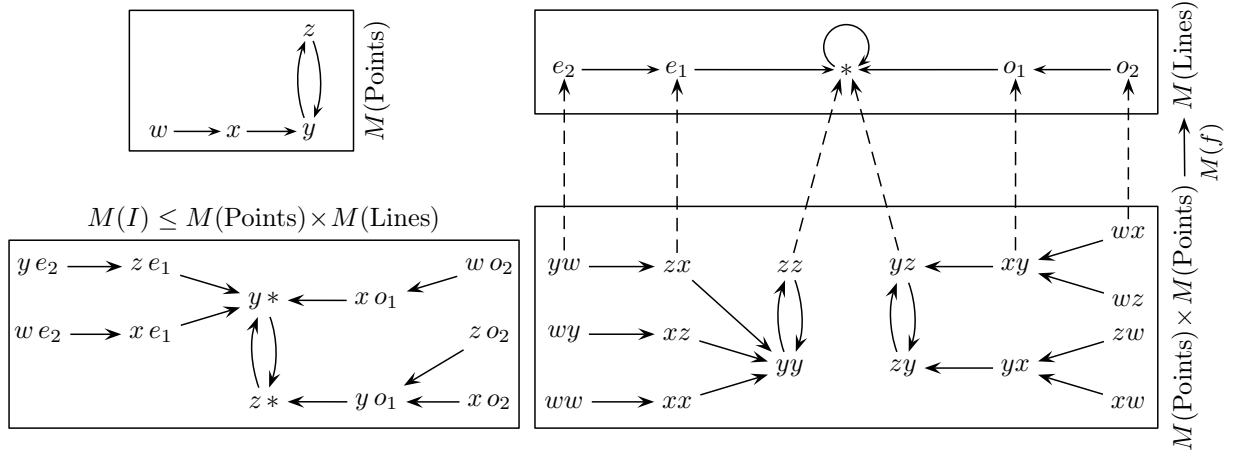
\newcommand{\Points}{
\begin{pspicture}(0,0)(2.4,1.8)
\psset{arrowsize=3pt 1, nodesep=2pt, linewidth=0.02, yunit=0.9}
\rput{90}(2.9,0.75){$M(\mbox{Points})$}
\psframe(-0.4,-0.3)(2.6,1.8)
\rput(0,0){\rnode{w1}{$w$}}
\rput(1,0){\rnode{x1}{$x$}}\ncline{->}{w1}{x1}
\rput(2,0){\rnode{y1}{$y$}}\ncline{->}{x1}{y1}
\rput(2,1.5){\rnode{z1}{$z$}}\ncarc[arcangle=20]{->}{y1}{z1}\ncarc[arcangle=20]{->}{z1}{y1}
\end{pspicture}}

\newcommand{\Lines}{
\begin{pspicture}(0,0)(7.5,0.9)
\psset{arrowsize=3pt 1, nodesep=2pt, linewidth=0.02}
\psframe(-0.4,-0.1)(7.9,1.3)
\rput(0,0.5){\rnode{e2}{$e_2$}}
\rput(1.5,0.5){\rnode{e1}{$e_1$}}\ncline{->}{e2}{e1}
\rput(3.75,0.5){\rnode{star}{$*$}}\ncline{->}{e1}{star}\nccircle{<-}{star}{0.3cm}
%
\rput(6,0.5){\rnode{o1}{$o_1$}}\ncline{->}{o1}{star}
\rput(7.5,0.5){\rnode{o2}{$o_2$}}\ncline{->}{o2}{o1}
\rput{90}(8.25,0.7){\pnode(0,0.1){End2}$M(\mbox{Lines})$}
\end{pspicture}
}

\newcommand{\PtimesP}{
\begin{pspicture}(0,0)(7.5,1)
\psset{arrowsize=4pt 1, nodesep=3pt, linewidth=0.02, yunit=0.9}
\psframe(-0.4,-0.4)(7.9,2.9)
\rput(0,0){\rnode{ww}{$ww$}}
\rput(0,1){\rnode{wy}{$wy$}}
\rput(0,2){\rnode{yw}{$yw$}}
\rput(1.5,0){\rnode{xx}{$xx$}}\ncline{->}{ww}{xx}
\rput(1.5,1){\rnode{xz}{$xz$}}\ncline{->}{wy}{xz}
\rput(1.5,2){\rnode{zx}{$zx$}}\ncline{->}{yw}{zx}
\rput(3,0.5){\rnode{yy}{$yy$}}\ncline{->}{xx}{yy}\ncline{->}{xz}{yy}\ncline{->}{zx}{yy}
\rput(3,2.0){\rnode{zz}{$zz$}}\ncarc[arcangle=20]{->}{yy}{zz}\ncarc[arcangle=20]{->}{zz}{yy}
\rput(7.5,0.0){\rnode{xw}{$xw$}}
\rput(7.5,1.0){\rnode{zw}{$zw$}}
\rput(7.5,1.5){\rnode{wz}{$wz$}}
\rput(7.5,2.5){\rnode{wx}{$wx$}}
\rput(6,0.5){\rnode{yx}{$yx$}}\ncline{->}{xw}{yx}\ncline{->}{zw}{yx}
\rput(6,2.0){\rnode{xy}{$xy$}}\ncline{->}{wx}{xy}\ncline{->}{wz}{xy}
\rput(4.5,0.5){\rnode{zy}{$zy$}}\ncline{->}{yx}{zy}\ncline{->}{xy}{yz}
\rput(4.5,2.0){\rnode{yz}{$yz$}}\ncarc[arcangle=20]{->}{zy}{yz}\ncarc[arcangle=20]{->}{yz}{zy}
\ncline{->}{xy}{yz}
\rput{90}(8.25,1.25){$M(\mbox{Points})\!\times\! M(\mbox{Points})$\pnode(0,0.1){End1}}
\rput(7.5,4.9){\rnode{wxup}{}}\ncline[linestyle=dashed]{->}{wx}{wxup}
\rput(6,4.9){\rnode{xyup}{}}  \ncline[linestyle=dashed]{->}{xy}{xyup}
\rput(3.8,4.9){\rnode{yzup}{}}  \ncline[linestyle=dashed]{->}{yz}{yzup}
\rput(3.7,4.9){\rnode{zzup}{}}  \ncline[linestyle=dashed]{->}{zz}{zzup}
\rput(1.5,4.9){\rnode{zxup}{}}  \ncline[linestyle=dashed]{->}{zx}{zxup}
\rput(0,4.9){\rnode{ywup}{}}    \ncline[linestyle=dashed]{->}{yw}{ywup}
\ncline{->}{End1}{End2}\nbput[nrot=90,npos=0.4]{\small $M(f)$}
\end{pspicture}
}

\newcommand{\Irel}{
\begin{pspicture}(0,0)(5.5,2.5)
\psset{arrowsize=4pt 1, nodesep=3pt, linewidth=0.02, yunit=0.9}
\psframe(-0.4,-0.4)(6.4,2.4)
\rput(0,2.0){\rnode{ye2}{$y\,e_2$}}
\rput(0,1.0){\rnode{we2}{$w\,e_2$}}
\rput(1.5,2.0){\rnode{ze1}{$z\,e_1$}}\ncline{->}{ye2}{ze1}
\rput(1.5,1.0){\rnode{xe1}{$x\,e_1$}}\ncline{->}{we2}{xe1}
%
%
\rput(6.0,2.0){\rnode{wo2}{$w\,o_2$}}
\rput(4.5,1.5){\rnode{xo1}{$x\,o_1$}}\ncline{->}{wo2}{xo1}
\rput(3.0,1.5){\rnode{ystar}{$y\,*$}}\ncline{->}{ze1}{ystar}\ncline{->}{xe1}{ystar}\ncline{->}{xo1}{ystar}
\rput(3.0,0.0){\rnode{zstar}{$z\,*$}}\ncarc[arcangle=20]{->}{zstar}{ystar}\ncarc[arcangle=20]{->}{ystar}{zstar}
\rput(4.5,0.0){\rnode{yo1}{$y\,o_1$}}
\rput(6,0){\rnode{xo2}{$x\,o_2$}}\ncline{->}{xo2}{yo1}
\rput(6.0,1.0){\rnode{zo2}{$z\,o_2$}}\ncline{->}{zo2}{yo1}\ncline{->}{yo1}{zstar}
\rput[b](3.0,2.5){$M(I)\leq M(\mbox{Points})\!\times\!M(\mbox{Lines})$}
\end{pspicture}}

\begin{center}
\begin{pspicture}(0,-0.0)(15,5.35)
\rput(2.8,4.8){\Points}
\rput(10.8,4.67){\Lines}
\rput[b](10.8,0.3){\PtimesP}
\rput[b](2.75,0.3){\Irel}
\end{pspicture}
\end{center}


There is a vast literature on categories whose objects are used in AI.
Categories of probability spaces can be traced  to \cite{Cencov, Lawvere161}.
A  sample of other resources includes \cite{BDEP, Giry, Jackson, Meng, Schiopu, Wendt}.
Note that use of semantic categories of probabilistic objects 
differs from assigning probabilities to logical formulae~\cite{Domingos, Scott, Fusion}.
Categories of fuzzy sets are well-studied~\cite{Eytan, Goguen, Elephant, Stout, Wyler}.
 Belief functions occur in~\cite{MDA, Woj08}.\vspace{-3pt}

\section{Categorical Logic}

\subsection{Syntax}
The set $\Sigma$-Type of first-order \textit{types} generated by a set $\Sigma$-Sort
of \textit{sorts}
consists of 
finite lists $A_1$, $\dots$, $A_n$ of sorts including the empty list
which is written $[\,]$. 
A first-order \textit{signature} $\Sigma$ has:
(1)~a set $\Sigma$-Sort of sorts;
(2)~a set $\Sigma$-Fun of \textit{function symbols} together with maps
$\Sigma\mbox{-Fun}\to\Sigma\mbox{-Type}$ and 
$\Sigma\mbox{-Fun}\to\Sigma\mbox{-Sort}$ 
respectively assigning to each function symbol
its \textit{type} and \textit{sort};
(3)~a set $\Sigma$-Rel of \textit{relation symbols} together with a map
$\Sigma\mbox{-Rel}\to\Sigma\mbox{-Type}$ assigning to each relation symbol its type;
(4)~a set $\Sigma$-Var of \textit{variables} together with a map
  $\Sigma\mbox{-Var}\to\Sigma\mbox{-Sort}$ 
assigning to each variable its \textit{sort}.
$f:A_1, \dots, A_n\to B$ indicates that $f$ 
is a function symbol with  type $A_1,\dots,A_n$ and sort~$B$.
If $n=0$, then $f$ is a \textit{constant}.
$R\rightarrowtail A_1,\dots,A_n$ indicates that $R$ is a relation symbol of 
type $A_1,\dots,A_n$.
If $n=0$, then $R$ is a \textit{proposition}.
$x:A$ indicates that $x$ is a variable of sort~$A$.
We assume a countably infinite supply of variables of each sort.

\subsubsection{Terms and Formulae}

\newcommand{\FV}{\mbox{FV}}
\newcommand{\fst}{\mbox{fst}}
\newcommand{\snd}{\mbox{snd}}
We recursively define the \textit{terms} 
over a  signature $\Sigma$ together with 
 the \textit{sort} $t:A$ and the set $\FV(t)$ 
of \textit{variables} of each term.
(1)~A variable $x:A$ is a term with $\FV(x)=\{x\}$. 
(2)~If $f:A_1,\dots,A_n\to B$ is a function symbol and $t_i:A_i$ are terms, 
then $f(t_1,\dots,t_n):B$ is a 
term with $\FV(f(t_1,\dots,t_n))=\bigcup\FV(t_i)$. 

We recursively define the \textit{formulae} over a signature $\Sigma$ together
with the set $\FV(\varphi)$ 
of \textit{free variables} of each formula.
 (1)~If $R\rightarrowtail A_1,\dots,A_n$ is a relation 
symbol and $t_i:A_i$ are terms, then $R(t_1,\dots,t_n)$ is a formula 
with $\FV(R(t_1,\dots,t_n))=\bigcup\FV(t_i)$.
(2)~If $s:A$ and $t:A$ are terms, then $(s=_At)$ is a formula with
$\FV(s=_At)=\FV(s)\cup \FV(t)$.
(3)~$\top$ and $\bot$ are formulae. Neither has free variables.
(4)~If $\varphi$ and $\psi$ are formulae and $\star$ is a symbol in $\{\wedge, \vee, \Rightarrow\}$,
   then  $\varphi\star\psi$ is a formula with $\FV(\varphi\star\psi)=\FV(\varphi)\cup\FV(\psi)$.
(5)~If $\varphi$ is a formula, then so is $\neg\varphi$ with $\FV(\neg\varphi)=\FV(\varphi)$.
(6)~If $\varphi$ is a formula, then $(\exists x:A)\,\varphi$ 
    and $(\forall x:A)\,\varphi$ are formulae.  
    Each has $\FV(\varphi)/\{x\}$ as its set of free variables.
Formulae constructed using (1)--(2) are \textit{atomic}.
Those built with $\top$, $\wedge$ and atomic formulae are \textit{Horn};
with $\exists$ and Horn are \textit{regular}; and with 
$\bot$, $\wedge$ and regular are \textit{coherent}.
All are \textit{first order}. A signature with no sorts is \textit{propositional}.

A \textit{context} is a finite list  $\vec{x}=x_1,\dots, x_n$ of distinct variables.
Its \textit{type} is $A_1,\cdots,\,A_n$ where $x_i:A_i$ and its \textit{length} is $n$.
The \textit{concatenation} of contexts $\vec{x}$ and $\vec{y}$ is  $\vec{x},\vec{y\,}'$
where $\vec{y\,}'=\{y\in\vec{y}\,|\,y\not\in\vec{x}\}$.
A context is \textit{suitable} for a term $t$ 
  if $\FV(t)\subset\vec{x}$.
If $t$ is a term and $\vec{x}$ is a context suitable for $t$, then  $\vec{x}.\,t$
is a \textit{term-in-context}. 
A context is \textit{suitable} for a formula $\varphi$ 
  if  $\FV(\varphi)\subset\vec{x}$. 
If $\varphi$ is a formula and $\vec{x}$ is a context suitable for $\varphi$, then  
 $\vec{x}.\varphi$ is a \textit{formula-in-context}. 
A context suitable for a term $t$ or formula $\varphi$ may include variables that do not occur
in $t$ or $\varphi$  (in addition to all those that do occur).
A context is \textit{suitable} for a list $\vec{s}$ of terms if it is suitable for 
each  $s_i\in\vec{s}$ and similarly for a list of formulae.
The \textit{canonical context} for a term $t$ or formula $\varphi$ consists of
the (free) variables of $t$ or $\varphi$ in  order of  occurrence.
We write  1 for the empty context.

\subsubsection{Substitution}

A \textit{substitution} $\theta=[\vec{s}\,/\, \vec{y}\,]$ consists of a 
context $\vec{y}$ and a list $\vec{s}$ of terms having the same length
and type as $\vec{y}$.    The empty substitution is $[\,]$.  
If $\vec{z}$ is a context, the \textit{extension} of $\theta$ to $\vec{z}$
is $\theta^{\,\vec{z}}=[\vec{s},\, \vec{z\,}'\,/\,\vec{y},\,\vec{z\,}']$ where
$\vec{z\,}'=\{z\in\vec{z}\;|\,z\not\in\vec{y}\}$.  For example, if
$\theta=[f(y),\,u\,/\,x,\,w]$ and $\vec{z}=w,z$  then $\theta^{\,\vec{z}}=[f(y),\,u,\,z\,/\,x,\,w,\,z]$.
A context is \textit{suitable} for a substitution if it is suitable for $\vec{s}$.
The \textit{canonical context} for a substitution $\theta=[\vec{s}\,/\, \vec{y}\,]$ 
is the canonical context for $\vec{s}$. 
Application of a substitution to a term is:\vspace{-3pt}
\[t[\vec{s}/\vec{y}\,] = 
  \left\{\begin{array}{ll}x &\mbox{if $t=x$ and $x\not\in \vec{y}$}\\[1pt] 
                          s_i & \mbox{if $t=y_i$ for some $y_i\in \vec{y}$}\\[1pt]
                          f(t_1[\vec{s}/\vec{y}\,],\,\dots,\,t_n[\vec{s}/\vec{y}\,])
                             &\mbox{if $t=f(t_1, \dots, t_n)$}\end{array}\right.\vspace{-3pt}\]
The  substitutions $[s_i/y_i]$ are performed simultaneously.  
For example,
$f(x, y)[x,z\,/\,y, x\,] = f(z,x)$.  In general, this differs from a sequential application:
$\big(f(x,y)[x\,/y\,]\big)[z\,/x\,] = f(x,x)[z\,/x\,] = f(z,z)$.
{\color{edit}We need not assume that $\vec{y}$ is suitable for $t$.  
If a variables of $t$ does not occur in $\vec{y}$, no substitution is applied to it.}
We apply a substitution $\theta=[\vec{s}/\vec{y}\,]$ to a formula $\varphi$
by simultaneously applying $\theta$ to all terms of $\varphi$ (D1.1.4 of~\cite{Elephant}).
In the case of quantified formulae, 
$\varphi[\vec{s}/\vec{y}\,]=(Q\,u')((\varphi_0\,[u'/u])\theta)$
             if $\varphi=(Q\,u)\varphi_0$ and $Q\in\{\exists,\,\forall\}$
where $u'$ is a variable of the same sort as $u$, $u'$ does not occur
   in $\vec{s}$ or $\vec{y}$, and $[u'/u]$ is applied to $\varphi_0$ before $\theta$ is applied.
For example, $((\exists x)\,R(x,y)\big)[x/y\,] = (\exists x')\,R(x', x)$.
Formulae are \textit{$\alpha$-equivalent} if they differ only in the names of their bound variables.
For example,
$((Q u)\varphi)$ and $((Q u')(\varphi [u'/u\,]))$ where $Q$ is a quantifier,
$u:U$, $u':U$ and $u'$ does not occur in $\varphi$.

First-order inference algorithms rely on unification~[\ref{closed-sorts}].
Unification constructs a substitution $\theta$ that, when applied to two lists 
of expressions, makes corresponding elements equal (or at least $\alpha$-equivalent).
Since we employ the sequent calculus of~\cite{Elephant}, unification must
apply to terms- and formulae-in-context rather than mere terms and formulae.  
Moreover, unification constructs $\theta$ in stages.  The fragment $\theta_i$ 
at stage $i$ is  built without awareness of the contexts occurring in later expressions.
We must, therefore, be able to apply a substitution $[\vec{s}/\vec{y}\,]$
to an expression-in-context $\vec{z}.e$ without $\vec{y}$ being suitable for $e$ and
without eliminating sorts that occur in $\vec{z}$.  

For a term-in-context,  define
$(\vec{z}.\,t)\theta = \vec{x}.(t\,\theta^{\,\vec{z}})$ where $\vec{x}$ is the canonical context
   for $\theta^{\,\vec{z}}$. For example:
$(x,y\,.f(x))[g(z),w\,/\,x,w]= (x,y\,.f(x))[g(z),w,y\,/\,x,w,y\,]$ $=
          z,w,y\,.\big(f(x)[g(z),w,y\,/\,x,w,y\,]\big) =z,w,y\,.f(g(z))$
Since $\vec{z}$ is suitable for $t$ and $\vec{x}$ is suitable for $\theta^{\vec{z}}$, 
$\vec{x}$ is suitable for $t\,\theta^{\vec{z}}$.
Moreover, if $z:Z$ is in $\vec{z}$, then either $z\in \vec{x}$ or there is a term
$s:Z$ in $(\vec{z}.\,t)[\vec{s}/\vec{y}\,]$.  
So, $(\vec{z}.\,t)\theta $ does not use the converse of weakening~[\ref{closed-sorts}].
\vfill
\eject

For a formula-in-context define:\vspace{-3pt}
\[(\vec{z}.\varphi)[\vec{s}/\vec{y}\,] = 
   \left\{\begin{array}{ll}
         \vec{x}.\varphi &\mbox{if $\varphi=\top$ or $\varphi=\bot$}\\[1pt]
         \vec{x}.(t_1\theta^{\,\vec{z}} = t_2\theta^{\,\vec{z}}) & \mbox{if $\varphi = (t_1=t_2)$}\\[1pt]
         \vec{x}.\,R(t_1\theta^{\,\vec{z}}\,,\dots,\, t_n\theta^{\,\vec{z}}) 
                   & \mbox{if $\varphi=R(t_1,\dots, t_n)$}\\[1pt]
         \vec{x}.((\varphi_0\theta^{\,\vec{z}}) * (\varphi_1\theta^{\,\vec{z}})) 
                       & \mbox{if $\varphi=(\varphi_0*\varphi_1)$
                               and $*\in\{\wedge, \vee, \Rightarrow\}$}\\[1pt]
         \vec{x}.(\neg(\varphi_0\theta^{\,\vec{z}})) & \mbox{if $\varphi=\neg\varphi_0$}\\[1pt]
         \vec{x}.\left((Q\,u')\,(\varphi_0[u'/u]\,\theta^{\vec{z}})\right)
                 & \mbox{if $\varphi=(Q\,u)\,\varphi_0$, $u'\not\in\vec{s},\vec{y},\vec{z}$,
     and $Q\in\{\exists, \forall\}$}\\
       \end{array}\right.\vspace{-3pt}\]
where $\vec{x}$ is the canonical context for $\theta^{\,\vec{z}}$. 
For example, 
$\big(y,z,w.((\exists x)\,R(x,y,z))\big)[f(x)/y]=$\\
    $\big(y,z,w.((\exists x')\,R(x',y,z)\big)[f(x), z, w/ y, z, w\,]
    = x,z,w.((\exists x')\, R(x', f(x), z)).$

\begin{lemma}
Lemma:  Let $\vec{z}.\,e_1$ and $\vec{z}.\,e_2$ be expressions-in-context with the same context
and let $\theta=[\vec{s}/\vec{y}\,]$ be a substitution.  Then $(\vec{z}.\,e_1)\,\theta$ and 
$(\vec{z}.\,e_2)\,\theta$ have the same context.
\end{lemma}

\myproof  $(\vec{z}.\,e_i)\,\theta = \vec{x}.(e_i\,\theta^{\vec{z}}\,)$ where $\vec{x}$ is the 
canonical context for $\theta^{\vec{z}}$ and is independent of $i$.\smallskip

\subsubsection{Deduction}\label{sec:deduction}
A \textit{sequent} is an expression $(\varphi\vdash_{\vec{x}}\psi)$ where
$\varphi$ and $\psi$ are formulae and $\vec{x}$ is a context suitable for 
both $\varphi$ and $\psi$. A \textit{theory} over a signature $\Sigma$ is a set of sequents.
A theory is classified as Horn, regular, coherent or intuitionistic according
  to the classification of formulae that occur in it.

For logical inference, we employ the sequent calculus of~\cite{Elephant}
shown  below.
 Rules  with a double-horizontal line may be used
in either direction. 
We assume that a variable that occurs bound in a sequent does not also occur free in that
sequent. 
The appendix (Section~[\ref{sec:appendix}]\,) includes proofs of derived rules.
Fragments of classical logic 
are obtained by including  different  connectives and their 
sequent rules.  Atomic logic 
 has the identity, cut, substitution and  $\mbox{Eq}0$.
Horn logic adds  $\mbox{Eq}1$ and the conjunction rules.  
Regular logic adds  $\exists$ and the Frobenius Axiom.  
Coherent logic adds the disjunction  and distributive rules.
Intuitionistic  logic adds  $\Rightarrow$ and $\forall$. Classical logic adds~EM. 
The distributive rule and Frobenius Axiom
are derivable in intuitionistic logic (see [\ref{thm:Frobenius}] and [\ref{thm:distrule}])
In regular logic we can derive the converse of Frobenius ([\ref{thm:Frobeniusconverse}]). 
In  coherent logic we can derive the converse of the distributive rule
[\ref{thm:distruleconverse}].


\begin{center}
{\setlength{\tabcolsep}{2pt}
\begin{tabular}{|rl|rl|}\hline
  Identity Rule (ID) & $\sequent{\varphi}{\vec{x}}{\varphi}$ & 
 Modus Ponens (Cut): \vphantom{\huge Y}&
$\frac{\sequent{\varphi}{\vec{x}}{\,\psi}\hspace{5pt}\sequent{\psi}{\vec{x}}{\,\chi}%
  \vphantom{\LARGE Y}}{%
    \sequent{\varphi}{\vec{x}}{\,\chi}}$
\vphantom{\large A}\\[4pt]\hline
Substitution (Sub):  \vphantom{\huge Y}&
   $\frac{\sequent{\varphi}{\vec{x}}{\psi}}{%
          \sequent{\varphi[\vec{s}/\vec{x}\,]}{\vec{y}}{\,\psi[\vec{s}/\vec{x}\,]}}$
&
Equality (Eq0): & $\sequent{\top}{x}{(x=x)}$\vphantom{\large Y}\\
\span 
     & (Eq1):& 
     $\sequent{((\vec{x}=\vec{y}\,)\wedge\varphi)}{\vec{z}}{\varphi[\vec{y}/\vec{x}\,]}$\\[3pt]\hline
%
True ($\top$):\vphantom{\huge Y}&
  $\sequent{\varphi}{\vec{x}}{\top}$ &
False ($\bot$): & $\sequent{\bot}{\vec{x}}{\varphi}$\vphantom{\large A}\\[3pt]
 And Elimination ($\wedge E$): &
   $\sequent{(\varphi\wedge \psi)}{\vec{x}}{\varphi}$ 
& Or Introduction ($\vee I$): & 
   $\sequent{\varphi}{\vec{x}}{(\varphi\vee \psi)}$\\[3pt]
& $\sequent{(\varphi\wedge \psi)}{\vec{x}}{\psi}$ & & 
   $\sequent{\psi}{\vec{x}}{(\varphi\vee \psi)}$\\[5pt]
And Rule ($\wedge$):& 
$\frac{\sequent{\varphi}{\vec{x}}{\,\psi}\hspace{5pt} \sequent{\varphi}{\vec{x}}{\,\chi}}{%
   \sequent{\varphi}{\vec{x}}{\,(\psi\wedge\chi)}}$ &
Or Rule ($\vee$):& 
$\frac{\sequent{\varphi}{\vec{x}}{\,\chi}\hspace{5pt} \sequent{\psi}{\vec{x}}{\,\chi}}{%
   \sequent{(\varphi\vee\psi)}{\vec{x}}{\,\chi}}$
\\[5pt]\hline
Implication ($\Rightarrow$):\vphantom{\huge Y} & 
  \rput[l](0,-0.05){$\frac{\underline{(\psi \vdash_{\vec{x}}\, (\varphi\Rightarrow \chi))}}{{((\varphi\wedge\psi)\,\vdash_{\vec{x}}\, \chi)}}$}
& Distributive Rule:\span \vphantom{\large A}\\
& & 
 $\sequent{(\varphi\wedge (\psi\vee \chi))}{\vec{x}}{%
    (\varphi\wedge \psi)\vee (\varphi\wedge \chi)}$\span\\[3pt]\hline
Existential  ($\exists$):\vphantom{\huge Y} &  
      \rput[l](0,-0.05){$\frac{\sequent{\varphi}{\vec{x},y}{\psi}}{\overline{\sequent{(\exists y)\varphi}{\vec{x}\,}{\psi}}}$}
  & Universal  ($\forall$): &  
      \rput[l](0,-0.05){$\frac{\sequent{\varphi}{\vec{x},y}{\psi}}{\overline{\sequent{\varphi}{\vec{x}}{\,\,(\forall y)\psi}}}$}\\
Quantification\hspace{15pt} &  & Quantification\hspace{15pt} &\\[2pt]\hline
Excluded Middle {\tiny (EM)}: \vphantom{\huge Y}& $\sequent{\top\!\!}{\vec{x}}{(\varphi\vee \neg \varphi)}$ & 
   Frobenius Axiom:&  $\sequent{(\varphi\wedge(\exists y)\psi)\!\!}{\vec{x}}{%
    (\exists y)(\varphi\wedge\psi)}$\vphantom{\large A}%
  \\[3pt]\hline
\end{tabular}}
\end{center}

\subsubsection{Sequents vs Formulae} 

\cite{Luger, Mitchell, RN}  use formulae not sequents to define
theories, deduction, and algorithms for logic-based agents.
Cut is defined using $\Rightarrow$ and contexts are replaced by 
universal quantifiers. To adapt the algorithms to less expressive
fragments of first-order logic, we use the  following. 

\begin{theorem} (See D1.1.5 of~\cite{Elephant}).
In intuitionistic logic,  $\sequent{\varphi}{\vec{x}}{\psi}$
and 
$\sequent{\top}{1}{((\forall \vec{x}\,)(\varphi\Rightarrow\psi))}$
 are provably equivalent.
Consequently, in
  intuitionistic logic, we can replace sequents by formulae.
\end{theorem}

\myproof

\begin{tabular}{lllclll}
Given a sequent:\span &          
        && Given an implication formula:\span\\
  1 & $\sequent{\varphi}{\vec{x}}{\psi}$                   & Hypothesis 
        && 1 & $\sequent{\top}{1}{(\forall \vec{x})(\varphi\Rightarrow \psi)}$ & Hypothesis\\[3pt]
  2 & $\sequent{(\varphi\wedge \top)}{\vec{x}}{\varphi}$   & $\wedge E0$
        && 2 & $\sequent{\top}{\vec{x}}{(\varphi\Rightarrow \psi)}$ & $\forall$ 1\\[3pt]
  3 & $\sequent{(\varphi\wedge \top)}{\vec{x}}{\psi}$       & Cut 2, 1
        && 3 & $\sequent{(\varphi\wedge \top)}{\vec{x}}{\psi}$       & $\Rightarrow$ 2\\[3pt]
  4 & $\sequent{\top}{\vec{x}}{(\varphi\Rightarrow \psi)}$ & $\Rightarrow$ 3
        && 4 & $\sequent{\varphi}{\vec{x}}{\top}$                    & $\top$\\[3pt]
  5 & $\sequent{\top}{1}{(\forall \vec{x})(\varphi\Rightarrow \psi)}$ & $\forall$ 4
        && 5 & $\sequent{\varphi}{\vec{x}}{\varphi}$                 & ID\\
&&        && 6 & $\sequent{\varphi}{\vec{x}}{(\varphi\wedge \top)}$    & $\wedge$ 4, 5\\[3pt]
&&        && 7 & $\sequent{\varphi}{\vec{x}}{\psi}$                    & Cut 6, 3\vspace{-10pt}
  \end{tabular}

\subsubsection{Horn Clauses vs Horn Sequents}\label{horn-clauses}

In \cite{Mitchell, RN}, a clause is defined to be a 
         disjunction of literals ($\varphi$ or $\neg \varphi$
          with $\varphi$ atomic). It is
     a Horn clause if  at most one is positive.
If exactly one is positive, we can reformulate the idea using Horn sequents.

\begin{theorem}
Given a sequent 
  {\color{black}$\sequent{\top}{\vec{x}}{(\neg \varphi_1\vee \cdots \vee \neg \varphi_n\vee \psi)}$}
    in which $\psi$ and each $\varphi_i$ is atomic, 
we can derive $\sequent{(\varphi_1\wedge \cdots \wedge \varphi_n)}{\vec{x}}{\psi}$
in intuitionistic logic.
\end{theorem}

\myproof

  \begin{tabular}{lll}
  1 & $\sequent{\top}{\vec{x}}{(\neg \varphi_1\vee \cdots \vee \neg \varphi_n\vee \psi)}$ & 
       Hypothesis\\[2pt] 
  2 & $\sequent{(\neg \varphi_1\vee \cdots \vee\neg \varphi_n)}{\vec{x}}{%
          \neg(\varphi_1\wedge\cdots \wedge \varphi_n)}$ &  Theorem [\ref{thm:demorganA}]\\[2pt]
  3 & $\sequent{(\neg \varphi_1\vee \cdots \vee\neg \varphi_n\vee \psi)}{\vec{x}}{%
          (\neg(\varphi_1\wedge\cdots \wedge \varphi_n)\vee \psi)}$ &  Theorem [\ref{thm:tworuleor}]\\[2pt]
  4 & $\sequent{\top}{\vec{x}}{\neg(\varphi_1\wedge\cdots \wedge \varphi_n)\vee \psi}$ & 
        Cut 1, 3\\[2pt]
  5 & $\sequent{\neg(\varphi_1\wedge\cdots \wedge \varphi_n)\vee \psi)}{\vec{x}}{%
        ((\varphi_1\wedge \cdots \wedge \varphi_n)\Rightarrow \psi)}$ 
      & Theorem  [\ref{thm:imp3}] (no EM)\\[2pt]
  6 & $\sequent{\top}{\vec{x}}{((\varphi_1\wedge \cdots \wedge \varphi_n)\Rightarrow \psi)}$
      & Cut 4, 5 \\[2pt]
  7 & $\sequent{(\top\wedge (\varphi_1\wedge \cdots \wedge \varphi_n))}{\vec{x}}{\psi}$
      &  $\Rightarrow$ 6\\[2pt]
  8 & \color{black}$\sequent{(\varphi_1\wedge \cdots \wedge \varphi_n)}{\vec{x}}{\psi}$
      &  Theorem  [\ref{thm:topwedgecor}]\\[2pt] 
  \end{tabular}\\
 If $\psi = \neg\psi'$ we obtain 
    $\sequent{(\varphi_1\wedge \cdots \wedge \varphi_n\wedge \psi)}{\vec{x}}{\bot}$
   or 
    $\sequent{(\varphi_1\wedge \cdots \wedge \varphi_n)}{\vec{x}}{\neg\psi'}$
   neither of which is  Horn. 

\subsubsection{Normal Form for Horn Theories}

\begin{theorem}\label{thm:hornnormal}
(See D.1.3.10 of \cite{Elephant}).  Any Horn sequent is provably equivalent to 
a  list of  sequents of the form\vspace{-2pt}
${\sequent{(\varphi_1\wedge \cdots \wedge \varphi_n)}{\vec{x}}{\psi}}$
  where each $\varphi_i$ and $\psi$ is either  atomic or
   $\top$. 
\end{theorem}

\myproof
   Given 
      $\sequent{(\varphi_1\wedge \cdots \wedge \varphi_n)}{\vec{x}}{%
             (\psi_1\wedge \cdots \wedge \psi_m)}$, 
   by $\wedge E$, we can derive $\sequent{(\psi_1\wedge\cdots \wedge \psi_m)}{\vec{x}}{\psi_k}$  
            for  $1\leq k\leq m$.
   Cut then yields
     $\sequent{(\varphi_1\wedge \cdots \wedge \varphi_n)}{\vec{x}}{\psi_k)}$.
   Conversely, if we have a list of sequents in normal form, we can combine them into
     one sequent  using $\wedge$.\hfill\enpr

Applying the proof above to each sequent in a Horn theory, we obtain a
Corollary:  Every Horn theory is provably equivalent to a Horn theory
in which every sequent is in normal form.
Algorithm~[\ref{horn-normal}] in the appendix implements Theorem [\ref{thm:hornnormal}].
The output sequents are distinct and all  have the same context $\vec{x}$.
We may convert a Horn theory to normal form by applying Algorithm~[\ref{horn-normal}] to 
its sequents and eliminating redundancies. 
See Algorithm~[\ref{horn-theory-normal}].  


\subsection{Semantics}

Signatures and theories can be assigned interpretations in suitable categories \cite{Elephant}.

\subsubsection{$\Sigma$-Structures}

Let $\Sigma$ be a signature.
A $\Sigma\mbox{\textit{-structure}}$  in  a category $\mathcal{C}$
with finite products consists of functions assigning (1)~an object $M(A)$
to each sort, (2)~a morphism $M(f):M(A_1)\times\cdots \times M(A_n)\to M(B)$
to each function symbol $f:A_1,\dots,\,A_n\to B$, and (3)~a subobject
$M(R)\longmapsto M(A_1)\times\cdots\times M(A_n)$ to each
relation symbol $R\rightarrowtail A_1,\dots, A_n$. In particular, 
the empty type $[]$ has $M([])=\mbox{the terminal object}$ and so a 
constant $f:[]\to B$ is interpreted as a \textit{point} $1\to M(B)$
and a proposition $R\rightarrowtail 1$ is a \textit{truth value} 
$M(R)\longmapsto 1$.\vspace{-8pt}

\subsubsection{Terms- and Formulae-in Context}

$\Sigma$-structures can be extended to terms- and formulae-in context 
(D.1.2.3 and D.1.2.7 of \cite{Elephant}).
Given $\vec{x}.\,t$  with $\vec{x}:X_1,\dots, X_m$ and $t:B$, then
$\interp{\vec{x}.\,t}:M(X_1)\times\cdots\times M(X_m)\to M(B)$ is a morphism.
If $\vec{x}.\varphi$ is a formula-in-context, then 
$\interp{\vec{x}.\varphi}_M\leq M(X_1)\times \cdots\times M(X_m)$ is a subobject.
Semantics of the  connectives are implemented via operations in suitable
categories (D1.2.6 of~\cite{Elephant}).  Although the algorithms in this paper are syntactic, 
 examples in [\ref{sec:introduction}] and [\ref{closed-sorts}] 
rely on the fact that $\wedge$ is computed as a pullback.
A sequent $\sigma=\sequent{\varphi}{\vec{x}}{\psi}$ is \textit{satisfied} in $M$
if $\interp{\vec{x}.\varphi}_M\leq \interp{\vec{x}.\psi}_M$. 
A theory $\mathbb{T}$ is \textit{satisfied} in $M$ if all its sequents are.
D.1.3.2 and D.1.4.11 of~\cite{Elephant} provide soundness and completeness theorems
for categorical semantics.\vspace{-8pt}

\subsubsection{Substitution}

Semantics of substitution into terms and terms-in-context are computed by 
composition while semantics of substitution into  formulae and formulae-in-context
are computed by pullback. The properties for terms and formulae
(D1.2.4 and D1.2.7 of \cite{Elephant}) are included in 
the appendix of this paper [\ref{sec:subprop}].
\begin{wrapfigure}[7]{R}{6.2cm}
\begin{pspicture}(0,0)(2.0,5.0)
\rput(3.3,3.65){$\xymatrix @C=10pt{
M(\vec{X}')\ar@{->}[d]|(.4){(\interp{\vec{x}\,'.\,s_1},\dots,\interp{\vec{x}'.\,s_n})}
       & M(\vec{X})\ar[l]_{\pi\,_{\vec{X}\,'}}\ar@{->}[dr]^{\pi\,_{\vec{Z}'}}\ar@{.>}[d]
 \\ 
M(\vec{Y}) & M(\vec{Y})\times M(\vec{Z}') \ar@{->}[r]
        \ar@{->}[l]
        \ar@{->}[d]^{\pi_{\vec{U}}}
      & M(\vec{Z}')\\
& M(\vec{U})\ar@{->}[r]^{\interp{\vec{u}.\,t}} & M(B)
}$}
\end{pspicture}
\end{wrapfigure}
\begin{theorem}
\textit{Substitution Property for Terms-in-Context}: If $\vec{z}.\,t$ is a term-in-context and
$\theta=[\vec{s}/\vec{y}\,]$ is a substitution, then
$\interp{(\vec{z}.\,t)\theta} = 
\interp{\vec{x}.(t\,\theta^{\vec{z}})}=\interp{\vec{u}.\,t}\circ\pi_U \circ
    \big(\left(\,\interp{\vec{x}\,'.\,s_1},\,\dots,\, \interp{\vec{x}\,'.\,s_n}\right)\circ\pi_{\vec{X}\,'},\;
       \pi_{\vec{Z}\,'}\big)$
where $\vec{z}\,'=\{z\in\vec{z}\,\vert\,z\not\in \vec{y}\}$, 
$\vec{u}$ and $\vec{x}\,'$ 
are  the canonical contexts for $t$ and $\vec{s}$,
and $\pi_U$ and $\pi_{Z'}$ are projections.
\end{theorem}

\myproof The definition of substitution justifies the first equality of 
 $\interp{(\vec{z}.\,t)[\vec{s}/\vec{y}\,]}=
\interp{\vec{x}.(t[\vec{s},\vec{z}\,'/\vec{y},\vec{z}\,'])}=$
$\interp{\vec{y},\vec{z}\,'.\,t}\circ \big(\interp{\vec{x}.s_1}, \dots, 
      \interp{\vec{x}.s_n}, \interp{\vec{x}.z_1'}, \dots, \interp{\vec{x}.z_k'}\big)$.
 The Substitution Property for Terms gives the second.
If $\vec{u}\subset \vec{y}, \vec{z}\,'$ is the canonical context for $t$, then
 the Weakening Property  (D1.2.4 of~\cite{Elephant}) implies 
$\interp{\vec{y},\vec{z}\,'.\,t}=\interp{\vec{u}.\,t}\circ \pi_{\vec{U}}$.
Similarly, if $\vec{x}\,'$ is  canonical  for $\vec{s}$, then
$(\interp{\vec{x}.\,s_1},\dots, \interp{\vec{x}.\,s_n}) = 
(\interp{\vec{x}\,'.\,s_1},\dots, \interp{\vec{x}\,'.\,s_n})\circ \pi_{\vec{X}\,'}$.
 $\interp{\vec{x}.\,z_j'\,}$ is a projection, hence,
$\interp{\vec{y},\vec{z}\,'.\,t}\circ \big(\interp{\vec{x}.s_1}, \dots, 
      \interp{\vec{x}.s_n}, \interp{\vec{x}.z_1'}, \dots, \interp{\vec{x}.z_k'}\big)$\\
$=\interp{\vec{u}.\,t}\circ\pi_U\circ\big(\left(\interp{\vec{x}'.\,s_1},\dots,
        \interp{\vec{x}'.s_n}\big)\circ\pi_{\vec{X}\,'},\,\pi_{\vec{Z}\,'}\right)$.
\hfill \enpr

\begin{theorem}
\textit{Substitution Property for Formulae-in-Context}:  If $\vec{z}.\varphi$ is a formula-in-context
and $\theta=[\vec{s}/\vec{y}\,]$ is a substitution,
then $\interp{(\vec{z}.\varphi)\theta}=\interp{\vec{x}.(\varphi\,\theta^{\vec{z}\,'})}$ is computed as
pullback where $\vec{u}$ is the canonical context for $\varphi$.\vspace{-12pt}
\[\xymatrix{
\interp{\vec{x}.(\varphi\,\theta^{\vec{z}\,'})}\ar@{|->}[d]\ar[rrr] &&&
    \interp{\vec{y},\vec{z}\,'.\varphi}\ar@{|->}[d]\ar[r] & \interp{\vec{u}.\varphi}\ar[d]\\
   M(\vec{X}) \ar[rrr]_(.45){\left(\interp{\vec{x}.s_1},\dots,\interp{\vec{x}.s_n}),\; \pi_{\vec{Z}'}\right)} &&& 
     M(\vec{Y})\times M(\vec{Z}\,') \ar[r]\ar[r]_(.65){\pi_{\vec{U}}} & M(\vec{U})}\vspace{-3pt}\]
\end{theorem}

\myproof This follows from the  Substitution Property for Formulae
(D1.2.7 of \cite{Elephant}).


\vfill
\eject
\section{Forward Chaining for Propositional Horn Theories} 

Algorithm~[\ref{pl-fc}] determines if a Horn sequent 
$\sigma=\sequent{(R_1\wedge\cdots\wedge R_k)}{1}{S}$ in normal form
 is derivable in a propositional Horn theory~$\mathbb{T}$.  
It adapts the formula-based algorithm of \cite{RN} 
by replacing Horn clauses 
with Horn sequents (see~[\ref{horn-clauses}]). 
It maintains a queue $\mathcal{Q}$ of proposition symbols  and makes successive
passes through the $\mathbb{T}$-axioms. 
Symbols $U$ that have occurred in $\mathcal{Q}$ are the right sides of
derived sequents $\sequent{(R_1\wedge\cdots\wedge R_k)}{1}{U}$.
Algorithm~[\ref{pl-fc}] and the formula-based 
algorithm of~\cite{RN} differ in the way the queue is initialized.  
When using sequents, there are two sources for the initial queue.
Each $R_i$ is in the initial queue since
$\sequent{(R_1\wedge\cdots\wedge R_k)}{1}{R_i}$ is derivable by $\wedge E$.
The second source of symbols in the initial queue is right sides
of $\mathbb{T}$-axioms of the form $\sequent{\top}{1}{U}$ since
we can apply cut to the sequent rule $\sequent{(R_1\wedge\cdots\wedge R_k)}{1}{\top}$ 
and the axiom.
In each pass, we have, in effect, derived 
$\sequent{(R_1\wedge\cdots \wedge R_k)}{1}{(\wedge\mathcal{Q})}$.
We then consider each axiom $\sequent{\psi}{1}{P}$ of $\mathbb{T}$
  and seek to apply $\wedge E$ to derive
$\sequent{(\wedge\mathcal{Q})}{1}{\psi}$. 
If this is possible,  we apply cut to derive $\sequent{(R_1\wedge\cdots\wedge R_k)}{1}{P}$.  
This adds $P$ to $\mathcal{Q}$. 
\vspace{-5pt}


\begin{algorithm}[h] 
\caption{Determine if  $\sigma=\sequent{(R_1\wedge\cdots \wedge R_k)}{1}{S}$ 
     is derivable in a propositional Horn theory 
    $\mathbb{T}$.}\label{pl-fc}
\begin{algorithmic}[1]
\Procedure{propositional-forward-chaining}{$\mathbb{T}$, $\sigma$}
\State $n_s\gets \mbox{the number of sequents $\sigma_i=\sequent{(P_1^i\wedge\cdots \wedge P_{n_i}^i)}{1}{Q_i}$
    in $\mathbb{T}$}$
\State $n_p\gets \mbox{the number of proposition symbols in $\sigma$ and in the axioms of $\mathbb{T}$}$
\State $\mbox{queue}\gets$ a queue containing $R_1$, $\dots$, $R_k$ and all $Q_i$
    for which $(P_1^i\wedge \cdots \wedge P_{n_i}^i)=\top$
\State $\mbox{inferred}\gets \mbox{an array of size $n_p$ with $\mbox{inferred[U]}=\mbox{false}$}$ 
      for all $U$
\State $\mbox{count}\gets$ an array of size $n_s$ with $\mbox{count[i]}=$ the number of
        proposition symbols on the left of  $\sigma_i$
\While{queue is not empty}
   \State $U\gets \mbox{pop}(\mbox{queue})$
   \If{$U=Q$} \Return true       
   \EndIf
   \If{$\mbox{inferred[U]}=\mbox{false}$}
      \State $\mbox{inferred[U]}\gets \mbox{true}$
      \For{$i\gets 0$ to  $n_s-1$}
         \If{$U$ occurs on the left side of of $\sigma_i$}
            \State $\mbox{count[i]}\gets \mbox{count[i]}-1$
            \If{$\mbox{count[i]}=0$} push $U$ onto queue
            \EndIf
         \EndIf
      \EndFor
   \EndIf
\EndWhile
\State \Return false
\EndProcedure
\end{algorithmic}
\end{algorithm}\vspace{-5pt}
Consider a theory with axioms \hspace{5pt}
\cnodeput(0,0.08){11}{\scriptsize 1}\hspace{11pt}$\sequent{(A\wedge B)}{1}{D}$
and \hspace{5pt}
\cnodeput(0,0.08){11}{\scriptsize 2}\hspace{11pt}$\sequent{(C\wedge D)}{1}{E}$.
The figure below shows how   Algorithm~[\ref{pl-fc}] yields
 a derivation of $\sequent{(A\wedge B \wedge C)}{1}{E}$.
  Each horizontal line is an entry into the while loop.
 The $\mathcal{Q}$ history is written as a conjunction.
 The underlined symbol is no longer in $\mathcal{Q}$. $U$ is the 
 popped symbol.   Count  indicates the sequent $\sigma_i$ 
  for which  count is decremented.  Inferred is set to true for the underlined symbol in $\sigma_i$.
The Derivation column indicates the derived sequents.\vspace{-5pt}
\begin{center}
\begin{tabular}{ccccl}
 $\mathcal{Q}$ History &  $U$ &  Count &  Derivation\\\hline
$\underbar{A}\wedge B\wedge C$ & $A$ \vphantom{\LARGE Y}
   & $\scalebox{0.9}{\cnodeput(0.2,0.08){1}{\scriptsize 1}}\hspace{15pt} 
       \sequent{(\underbar{A}\wedge B)}{1}{D}$\\[2pt]\hline
$\underbar{A}\wedge\underbar{B}\wedge C$ & $B$ \vphantom{\LARGE Y}
   & $\scalebox{0.9}{\cnodeput(0.2,0.08){1}{\scriptsize 1}}\hspace{15pt} 
        \sequent{(\underbar{A}\wedge \underbar{B})}{1}{D}$
      & 1. $\sequent{(A\wedge B\wedge C)}{1}{(A\wedge B)}$\hfill\ & $\wedge$ E\\
  & & & 2. $\sequent{(A\wedge B)}{1}{D}$\hfill\ & Axiom 1\\
  & & & 3. $\sequent{(A\wedge B\wedge C)}{1}{D}$\hfill\ & Cut 1, 2\\[2pt]\hline
$\underbar{A}\wedge\underbar{B}\wedge\underbar{C}\wedge D$  \vphantom{\LARGE Y}
  & $C$ 
   & $\scalebox{0.9}{\cnodeput(0.2,0.08){1}{\scriptsize 2}}\hspace{15pt} 
        \sequent{(\underbar{C}\wedge D)}{1}{E}$
      & 4. $\sequent{(A\wedge B\wedge C)}{1}{(A\wedge B\wedge C)}$\hfill\ & Id \\    
&&&    5. $\sequent{(A\wedge B\wedge C)}{1}{(A\wedge B\wedge C\wedge D)}$ & $\wedge I$ 4, 3\\[2pt]\hline
$\underbar{A}\wedge\underbar{B}\wedge\underbar{C}\wedge \underbar{D}$
        \vphantom{\LARGE Y}%
  & $D$ 
   & $\scalebox{0.9}{\cnodeput(0.2,0.08){1}{\scriptsize 2}}\hspace{15pt} 
        \sequent{(\underbar{C}\wedge \underbar{D})}{1}{E}$
   & 6. $\sequent{(A\wedge B\wedge C \wedge D)}{1}{(C\wedge D)}$\hfill\ & $\wedge$E\\
&&& 7. $\sequent{(C\wedge D)}{1}{E}$\hfill\ & Axiom 2 \\
&&& 8. $\sequent{(A\wedge B\wedge C\wedge D)}{1}{E}$\hfill\ & Cut 6, 7\\
&&& 9. $\sequent{(A\wedge B\wedge C)}{1}{E}$\hfill\ & Cut 5, 8\\\hline
$\underbar{A}\wedge\underbar{B}\wedge\underbar{C}\wedge \underbar{D}\wedge \underbar{E}$ \vphantom{\huge Y}
   & $E$\\ 
\hphantom{} & \hphantom{D} & \hphantom{$\scalebox{0.9}{\cnodeput(0.2,0.08){1}{\scriptsize 2}}\hspace{15pt} 
        \sequent{(\underbar{C}\wedge \underbar{D})}{1}{E}$} & \\[-30pt]
\hphantom{Axiom 1}
\end{tabular}\vspace{-.5in}
\end{center}
\vfill
\eject

\section{Unification of Terms- and Formulae-in Context}\label{sec:unification}

A \textit{unification} of lists 
$[\vec{x}_1.\,\alpha_1,\dots, \vec{x}_n.\,\alpha_n]$ 
and $[\vec{y}_1.\,\beta_1,\dots, \vec{y}_n.\,\beta_n]$
of terms-in-context is a substitution $\theta$ for which 
$(\vec{x}_i.\alpha_i)\theta = (\vec{y}_i.\beta_i)\theta$ for $1\leq i\leq n$.  
For example, $\theta=[u,b,k,a,f(g(z)), g(z)\, /\, u,b,k,a,x,y\,]$
unifies $[u,x.\,g(x),\; a,y.\,f(y)]$ and $[b,x,y.\,g(f(y)),\; k,z.\,f(g(z))]$ since
\begin{center}
{\setlength{\tabcolsep}{2pt}
\begin{tabular}{rclcl}
$(u,x.\,g(x))\theta$     & $=$& $u,b,k,a,z.\,(g(x)\theta)$    & $=$ & $ u,b,k,a,z.\,g(f(g(z)))$\\
$(b,x,y.\,g(f(y))\theta$ & $=$& $u,b,k,a,z.\,(g(f(y))\theta)$ & $=$ & $ u,b,k,a,z.\,g(f(g(z)))$\\[3pt]
$(a,y.\,f(y)\theta$      & $=$& $u,b,k,a,z.\,(f(y)\theta)$    & $=$ & $u,b,k,a,z.\,f(g(z))$\\
$(k,z.\,f(g(z))\theta$   & $=$& $u,b,k,a,z.\,(f(g(z))\theta^z)$ & $=$ & $u,b,k,a,z.\,f(g(z))$
\end{tabular}}
\end{center}

A unification of lists of formulae-in-context is defined similarly.
Unification is an essential subroutine  for inference using fragments of first-order logic.  
We adapt the procedures of \cite{RN} and \cite{BPT} to support (1)~multi-sorted signatures
and (2)~terms- and formulae-in-context rather than terms and formulae without a context.
We must take contexts into account in order to correctly apply the substitution rule.
Unification algorithms taking into account only~(1) are included in~[\ref{unify1}]
and~[\ref{unify2}].  
Algorithms~[\ref{term-sub}], [\ref{simsub}], [\ref{alg:sub-term-in-context}], 
[\ref{formula-sub}],  [\ref{simsubphi}], 
and~[\ref{alg:sub-formula-in-context}]
of the appendix 
apply substitutions to terms, terms-in-context, formulae and 
formulae-in-context.

\begin{theorem}\label{u-works}
If Algorithm~[\ref{unification-terms-in-context}] returns a substitution $\theta$, 
then $(\vec{x}_i.\,\alpha_i)\,\theta = (\vec{y}_i.\,\beta_i)\theta$ for $1\leq i\leq~n$.
\end{theorem}

\noindent
\myproof
First note that  $(\vec{x}_1.\,\alpha_1)\theta_1 = (\vec{y}_1.\,\beta_1)\theta_1$:

Case A:  Applying $\theta_1$  concatenates $\vec{x}_1$ and 
$\vec{y}_1$ without $x$ then transforms both terms-in-context to $\vec{z},x.x$.  

Case B:  Applying $\theta_1$ concatenates $\vec{x}_1$ and $\vec{y}_1$ without $x$ or $y$
   then transforms both terms to $\vec{z},y.y$.

Case C: 
The terms-in-context are $\vec{x}_1.\,x$ and $\vec{y}_1.\,g(\,\vec{t}\,)$.  $\vec{z}$ consists of all 
variables of $\vec{x}_1$ and $\vec{y}_1$ except $x$ and $\mbox{FV}(\,\vec{t}\,)$.
  $\theta_1=[\vec{z}, g(\,\vec{t}\,) / \vec{z}, x\,]$.
Since
$(\vec{x}_1.\,x)\theta_1 =\vec{z},\mbox{FV}(\,\vec{t}\,).(x[\vec{z},g(\,\vec{t}\,),\mbox{FV}(\,\vec{t}\,)/\vec{z},x,\mbox{FV}(\,\vec{t}\,)]) 
   = \vec{z},\mbox{FV}(\,\vec{t}\,).\,g(\vec{t})$
and
$(\vec{y}_1.\,g(\,\vec{t}\,))\theta_1 =\vec{z},\mbox{FV}(\,\vec{t}\,).(g(\,\vec{t}\,)[\vec{z},g(\,\vec{t}\,),\mbox{FV}(\,\vec{t}\,)/\vec{z},x,\mbox{FV}(\,\vec{t}\,)]) = \vec{z},\mbox{FV}(\,\vec{t}\,).\,g(\,\vec{t}\,)$, we have
 $(\vec{x}_1.\alpha_1)\,\theta_1 = (\vec{y}_1.\beta_1)\,\theta_1$.

Case D:  
The procedure recursively calls itself with $[\vec{y}_1.\beta_1,\,\dots]$ and
  [$\vec{x}_1.\alpha_1,\, \dots]$  without removing terms or applying a substitution.

Case E: If the function symbols agree, the procedure recursively calls itself 
without removing terms or applying a substitution.

Consequently, $(\vec{x}_1.\,\alpha_1)\theta = (\vec{y}_1.\,\beta_1)\theta$.  
For each case A--E, if 
a substitution $\theta_i$ is appended to $\theta$, then it is applied to all subsequent 
terms-in-context in the recursive call.  Hence, when $\vec{x}_i.\,\alpha_i$ and 
$\vec{y}_i.\,\beta_i$ appear in the first terms in the argument lists, the algorithm
seeks to unify
$(\vec{x}_i.\,\alpha_i)\theta_1\cdots \theta_\ell$ and
$(\vec{y}_i.\,\beta_i)\theta_1\cdots \theta_\ell$ for some $\ell\geq 0$.
By induction, if $\theta_*$ is appended to $\theta$, we have
$(\vec{x}_i.\,\alpha_i)\theta_1\cdots \theta_\ell\,\theta_* = 
(\vec{y}_i.\,\beta_i)\theta_1\cdots \theta_\ell\,\theta_*$. 
It follows that 
$(\vec{x}_i.\,\alpha_i)\,\theta = 
(\vec{y}_i.\,\beta_i)\,\theta$. \hfill\enpr
\begin{algorithm}[h] 
\caption{Find a unification $\theta$ of two lists of terms-in-context}\label{unification-terms-in-context}
\begin{algorithmic}[1]
\Procedure{unify-terms-in-context}{$[\vec{x}_1.\alpha_1, \dots, \vec{x}_m.\alpha_m],\, 
      [\vec{y}_1.\beta_1, \dots, \vec{y}_n.\beta_n]$, $\theta=[]$}
\If{$m\not=n$} \Return Null
\ElsIf{$m = 0$} \Return $\theta$
\ElsIf{$\mbox{sort}(\alpha_1)\not=\mbox{sort}(\beta_1)$} \Return Null
\ElsIf{$\alpha = x:X$ is a variable}
  \If{$\beta_1=x$}\Comment Case A\label{Aline}
        \State $\vec{z}\gets \mbox{union}(\vec{x\,}_1, \vec{y\,}_1)/\{x\}$
        \State $\theta_1\gets [\vec{z},\,x\, /\, \vec{z},\,x]$
        \State \Return \Call{unify-terms-in-context}{$[(\vec{x}_2.\alpha_2)[\theta_1], \dots, ],\;
                                      [(\vec{y}_2.\beta_2)[\theta_1], \dots ],\; \theta\theta_1$}
  \ElsIf{$\beta_1=y\not=x$} \Comment Case B\label{Bline}
        \State $\vec{z}\gets \mbox{union}(\vec{x}_1,\,\vec{y\,}_1)/\{x, y\}$
        \State $\theta_1\gets [\vec{z},\,y\, /\, \vec{z},\,x]$          
       \State \Return \Call{unify-terms-in-context}{$[(\vec{x}_2.\alpha_2)[\theta_1], \dots, ],\;
                                         [(\vec{y}_2.\beta_2)[\theta_1], \dots ],\; \theta\theta_1$}
  \Else {\ $\beta_1 = g(t_1, \dots, t_\ell)$} \Comment Case C\label{Cline}
     \If{$\alpha_1=x$ occurs in $\beta_1$} \Return Null
     \Else{}
     \State $\vec{z}\gets \mbox{union}(\vec{x}_1,\,\vec{y}_1)/\{x\;\mbox{and the canonical 
               context of $\vec{t}$\,}\}$
     \State $\theta_1\gets [\vec{z},\, g(\vec{t})\, /\, \vec{z},\, x]$
      \State \Return \Call{unify-terms-in-context}{$[(\vec{x}_2.\alpha_2)[\theta_1], \dots, ],\;
                                         [(\vec{y}_2.\beta_2)[\theta_1], \dots ],\; \theta\theta_1$} 
     \EndIf
  \EndIf
\Else {\ $\alpha_1 = f(s_1, \dots, s_k)$}
   \If{$\beta_1$ is a variable} \Comment Case D\label{Dline}
          \State \Return \Call{unify-terms-in-context}{$[\vec{y}_1.\beta_1,\, \vec{x}_2.\alpha_2, \dots],\,
                  [\vec{x}_1.\alpha_1,\, \vec{y}_2.\beta_2, \dots],\,\theta$}
   \Else {\ $\beta_1 = g(t_1, \dots, t_\ell)$} \Comment Case E\label{Eline}
      \If{$f\not= g$} \Return Null
      \Else{}  
       \State \Return \Call{unify-terms-in-context}{$[\vec{x}_1.\,s_1, \dots, 
                             \vec{x}_2.\alpha_2, \dots],\,
                                       [\vec{y}_1.\,t_1, \dots, 
                                          \vec{y}_2.\beta_2, \dots],\,\theta$}
      \EndIf
    \EndIf
\EndIf
\EndProcedure
\end{algorithmic}
\end{algorithm}

\begin{lemma}\label{ut12}
If Algorithm~[\ref{unification-terms-in-context}] returns a substitution $\theta$, 
then $(\vec{x}_1.\,\alpha_1)\theta$ and $(\vec{x}_2.\,\alpha_2)\theta$ have the same 
context.
\end{lemma}

\myproof If $u$ is a variable in the context of $(\vec{x}_1.\alpha_1)\,\theta$ but not 
$(\vec{x}_2.\alpha_2)\,\theta$, then there is a minimum $j$ for which
$u\in (\vec{x}_1.\alpha_1)\,\theta_1\cdots\theta_j$ but 
$u\not\in (\vec{x}_2.\alpha_2)\,\theta_1\cdots\theta_j$
where $\theta=\theta_1\cdots\theta_k$.
If $1\leq j\leq k$, then 
$(\vec{x}_1.\alpha_1)\,\theta_1\cdots\theta_{j-1}$ and
$(\vec{x}_2.\alpha_2)\,\theta_1\cdots\theta_{j-1}$ both contain $u$ or neither does.
But $\theta_j$ will either add $u$ to both contexts or to neither.
If $j=0$, then $u\in (\vec{x_1}.\alpha_1)$ but $u\not\in (\vec{x_2}.\alpha_2)$.
If $\theta_1$ is generated by 
Case A then $\theta_1$ does not remove any variables, hence $u\in (\vec{x_2}.\alpha_2)$.
Cases B and C: If $\theta_1$ adds $u$ to contexts, then  $u\in (\vec{x_2}.\alpha_2)$.
  If $\theta_1$ substitutes some $y$ for $u$, then $u\not\in (\vec{x_1}.\alpha_1)\theta_1$.
Cases D and E:  No substitution is applied.  
\hfill \enpr

\begin{lemma}\label{utij}
If Algorithm~[\ref{unification-terms-in-context}] returns  $\theta$, 
then $(\vec{x}_i.\,\alpha_i)\theta$ and $(\vec{x}_j.\,\alpha_j)\theta$ have the same 
context for all $i$, $j$.
\end{lemma}

\myproof We may assume without loss of generality that $i< j$.
As an intermediate step in the procedure we will reach a recursive call to unify 
$[(\vec{x}_i.\,\alpha_i)\theta_*,\; (\vec{x}_{i+1}.\,\alpha_{i+1})\theta_*,\; \dots]$ and
$[(\vec{y}_i.\,\beta_i)\theta_*,\; (\vec{y}_{i+1}.\,\beta_{i+1})\theta_*,\; \dots]$ where
$\theta_*$ is the part of $\theta$ constructed so far. Since we assume the algorithm
returns a substitution, the algorithm will proceed to construct the remaining part
$\theta'$ of $\theta=\theta_*\,\theta'$.
By Lemma~[\ref{ut12}], 
$(\vec{x}_i.\,\alpha_i)\theta_*\,\theta'$ and $(\vec{x}_{i+1}.\,\alpha_{i+1})\theta_*\theta'$ have the same 
context.  The result follows by induction. \hfill\enpr

\begin{theorem}\label{u-context}
If Algorithm~[\ref{unification-terms-in-context}] returns  $\theta$, 
then $(\vec{x}_i.\alpha_i)\,\theta$ and $(\vec{y}_j.\beta_j)\,\theta$ have the same
context for all $i$, $j$.
\end{theorem}

\myproof By Lemma~[\ref{utij}], all $(\vec{x}_i.\,\alpha_i)\,\theta$ have the
same context.  By Theorem~[\ref{u-works}], 
$(\vec{x}_i.\,\alpha_i)\,\theta = (\vec{y}_i.\,\beta_i)\,\theta$, hence, they have
the same context.  \hfill\enpr
\medskip

\noindent
Consider, for example:
$\mbox{\textsc{unify-terms-in-context}}([x,u.\,g(x),\, a,y.f(y)],\; [b,x,y.\,g(f(y)),\,k,z.f(g(z))])$.
Case E results in
$\mbox{\textsc{unify-terms-in-context}}([x,u.\,x,\;\; a,y.\,f(y)], [b,x,y.\,f(y),\;\; k,z.\,f(g(z))])$.
 Case C \\results in $\theta_1=[u,b,f(y)\,/\,u,b,x\,]$.
  Next call 
   $\mbox{\textsc{unify-terms-in-context}}([(a,y.f(y))\theta_1],\; [(k,z.f(g(z)))\theta_1])$. \\
Evaluating the arguments yields:
$(a,y.f(y))\theta_1=a,b,y,a.\left(f(y)[u,b,f(y),a,y\,/\,a,b,x,a,y]\right) = u,b,y,a.\,f(y)$ and
$(k,z.f(g(z)))\theta_1 = u,b,y,k,z.\left(f(g(z))[u,b,f(y),k,z\,/\,u,b,x,k,z]\right)=
    u,b,y,k,z.\,f(g(z))$. 
 Case E\\ results in 
$\mbox{\textsc{unify-terms-in-context}}([u,b,y,a.\,y\,],\;[u,b,y,k,z.\,g(z)]$.
 Case C then yields \\$\theta_2=[u,b,k,a,g(z)\;/\; u,b,k,a,y\,]$.
 The resulting substitution is $\theta_1\,\theta_2=[u,b,k,a,f(g(z)),g(z)\;/\;u,b,k,a,x,y\,]$.

\begin{algorithm}[h] 
\caption{Find a unification $\theta$ of two lists of formulae-in-context.}\label{unification-fic}
\begin{algorithmic}[1]
\Procedure{unify-formulae-in-context}{$[\vec{x}_1.\alpha_1, \dots, \vec{x_m}.\alpha_m],\,
           [\vec{y_1}.\beta_1,\dots, \vec{y}_m\beta_n], \theta=[])$}
\If{$m\not= n$} \Return Null
\ElsIf{$m=0$} \Return $\theta$
\ElsIf{$\alpha_1=(s_1 = s_2)$ and $\beta_1=(t_1 = t_2)$} 
  \State $\theta' = \mbox{\Call{unify-terms-in-context}{$[\vec{x}_1.\,s_1,\; \vec{x}_1.\,s_2],\; 
        [\vec{y}_1.\,t_1, \vec{y}_1.\,t_2],\,\theta$}}$
  \State \Return \Call{unify-formulae-in-context}{$[(\vec{x}_2.\,\alpha_2)\,\theta', \dots],\,
      [(\vec{y}_2.\,\beta_2)\,\theta',\dots],\,\theta\,\theta'$}
\ElsIf {$\alpha_1=Q(s_1, \dots, s_k)$ and $\beta_1=R(t_1, \dots, t_\ell)$}
   \If {$Q\not=R$} \Return Null
   \Else
      \State $\theta'=\mbox{\Call{unify-terms-in-context}{$[\vec{x}_1.\,s_1, \dots, \vec{x}_1.s_k],\;
          [\vec{y}_1.\,t_1, \dots, \vec{y}_1.\,t_k], \theta$}}$
      \State \Return \Call{unify-formulae-in-context}{$[(\vec{x}_2.\,\alpha_2)\theta', \dots, ],\;
         [(\vec{y}_2.\beta_2)\,\theta',\dots], \theta\,\theta'$}
    \EndIf
\ElsIf {$\alpha_1 = \varphi*\varphi'$ and $\beta_1=\psi * \psi'$ with $*=\wedge$, $\vee$ or $\Rightarrow$}
   \State \Return \Call{unify-formulae-in-context}{%
             $[\vec{x_1}.\,\varphi, \vec{x_1}.\,\varphi', \vec{x_2}.\,\alpha_2, \dots],\, 
                          [\vec{y_1}.\psi,\,\vec{y_1}.\,\psi', \vec{y_2}.\beta_2,\dots],\, \theta$}
\ElsIf{$\alpha_1=\neg\varphi$ and $\beta_1=\neg\psi$}
   \State \Return \Call{unify-formulae-in-context}{$[\vec{x_1}.\,\varphi,\, \vec{x_2}.\alpha_2,\, \dots],\; 
                          [\vec{y_1}.\,\psi,\, \vec{y_2}.\beta_2,\dots],\, \theta$}
\ElsIf{$\alpha_1=((Q.x)\varphi)$ and $\beta_1=((Q.y)\psi)$ with $Q=\exists$ or $\forall$}
   \If{$\mbox{sort}(x)\not=\mbox{sort}(y)$} \Return Null
   \Else{} 
     \State Let $u_1$, $u_2$,  and $u_3$ be variables of $\mbox{sort}(x)$ distinct from 
            those in $\vec{x}_1$, $\vec{y}_1$, $x$, $y$, and each other.
     \If{$x=y$}  
            \State let $u_2=u_3$. 
     \EndIf
            \State \Return 
           \textsc{unify-formulae-in-context}$($ 
           \State \hphantom{return \ \,\textsc{unify}}
                  $[(\vec{x}_1,u_1.\varphi[u_1/x])[u_3/y\,],\;(\vec{x}_2.\alpha_2)[u_2, u_3/x, y\,], \dots],\;$
           \State\hphantom{return \ \,\textsc{unify}}  $[(\vec{y_1},u_1.\psi[u_1/y])[u_2/x],\;\,
                       (\vec{y_2}.\beta_2)[u_2, u_3/x, y], \dots],\;
                                                 \theta[u_2, u_3 / x, y])$
   \EndIf
\EndIf
\State \Return Null
\EndProcedure
\end{algorithmic}
\end{algorithm}

\vfill
\eject

\section{Detecting Closed Sorts}\label{closed-sorts}

\newcommand{\three}{\begin{pspicture}(0,0.2)(1,0.8)
\psset{arrowsize=3pt 2, yunit=0.85}
   \dotnode[dotstyle=*](0.5,1){A1}
   \dotnode[dotstyle=*](0,0){A2}  \ncline{->}{A1}{A2}
   \dotnode[dotstyle=*](1.0,0){A3}\ncline{->}{A2}{A3}\ncline{->}{A3}{A1}
  \end{pspicture}}
\newcommand{\two}{\begin{pspicture}(0,0)(0.5,1)
\psset{arrowsize=3pt 2}
   \dotnode[dotstyle=*](0,0){B1}
   \dotnode[dotstyle=*](0,1){B2}\ncarc[arcangle=25]{->}{B1}{B2}\ncarc[arcangle=25]{->}{B2}{B1}
  \end{pspicture}}
\begin{wrapfigure}[10]{R}{6.4cm}
\begin{center}
\begin{pspicture}(0.3,-2.0)(9.5,1.7)
\psset{yunit=0.9}
\psset{arrowsize=3pt 2}
\rput[l](0.8,2){$M(A) = $}  \rput(2.2,1.2){\psframe(0,0)(2.5,1.6)}\rput(3,2){\three}\rput(4.5,2){\two}
\rput[l](0.4,-0.8){$M(R) = $}\rput(1.8,-1.5){\psframe(0,0)(1.4,1.4)}\rput(2.5,-0.7){\three}
   \pnode(2.5,0.2){X1}
   \pnode(2.8,1.0){X2}\ncline{|->}{X1}{X2}
\rput(4.4,-1.5){\psframe(0,0)(0.7,1.5)}\rput(5.0,-0.7){\two}
   \rput[l](5.3,-0.8){$=M(S)$}
   \pnode(4.6,0.3){X3}
   \pnode(4.3,1.0){X4}\ncline{|->}{X3}{X4}
\rput(6.0,2){$M(B)=\phi$}
\end{pspicture}
\end{center}
\end{wrapfigure}
The substitution rule includes the weakening rule: from $\sequent{\varphi}{\vec{x}}{\psi}$
derive $\sequent{\varphi}{\vec{y}}{\psi}$
where  $\vec{y}$ contains all the variables of $\vec{x}$ and possibly others. 
That is, we are free to introduce new variables into a context.
Since $\vec{x}$ must be suitable for $\varphi$ and $\psi$, no new variable of $\vec{y}$ 
is free in either formula.
The converse of weakening is not a permitted inference rule. 
There are two exceptions. 
Let $y:A$ be a variable that occurs in $\vec{y}$ but not  $\vec{x}$.
(1)~If $x:A$  occurs in $\vec{x}$, then we may substitute
$[x/y\,]$ into $\sequent{\varphi}{\vec{y}}{\psi}$.
(2)~If there is a closed term $k:Y$, then
we may substitute $[k/y\,]$ into  $\sequent{\varphi}{\vec{y}}{\psi}$.
In either case, we eliminate $y$ from $\vec{y}$.

As a counterexample to the converse of weakening, 
let $\mathbb{T}$ be the Horn theory 
with sorts $A$ and $B$, relations 
$S\rightarrowtail A$ and $R\rightarrowtail A$ and
axiom $\sequent{R(a)}{a,b}{S(a)}$ where $a:A$ and $b:B$. 
Construct a model $M$ of $\mathbb{T}$ in $\Set^{\circlearrowleft}$ as shown above.
$\interpretation{a,b.R(a)}$ and $\interpretation{a,b.S(a)}$
are subobjects of $M(A)\times M(B) = \phi$, hence,  both equal $\phi$.  
Consequently, $M\models \sequent{R(a)}{a,b}{S(a)}$.  However,
$\interpretation{a.R(a)}\cong M(R)$ and $\interpretation{a.S(a)}\cong M(S)$ 
in $M(A)$.  Since there is no $\Set^{\circlearrowleft}$ morphism from a 3-cycle to a 2-cycle,
$M\not\models \sequent{R(a)}{a}{S(a)}$.  If $\sequent{R(a)}{a}{S(a)}$ were derivable in 
$\mathbb{T}$ this would contradict the Soundness Theorem (Proposition D.1.3.2 of \cite{Elephant}).


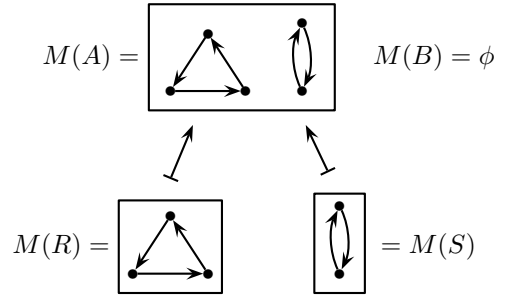
\vfill
\eject

An application of the  special cases of the converse to weakening arises in 
Algorithm~[\ref{fo-fc}].  
If $y:A$ is in $\vec{y}$, it is 
simple to determine if there is an $x:A$ in $\vec{x}$.  
Algorithm [\ref{closed-sort}] determines if there is a closed term $k:A$.

\begin{algorithm}[h] 
\caption{Determine if a sort $A$ of a signature $\Sigma$ is closed.}\label{closed-sort}
\begin{algorithmic}[1]
\Procedure{sort-closed?}{$\Sigma$, $A$}
\State $n_s\gets \mbox{the number of sorts in $\Sigma$}$
\State $n_f\gets \mbox{the number of function symbols in $\Sigma$}$
\State $\mbox{queue}\gets \mbox{a queue containing the sorts $S$ of $\Sigma$ for which 
        $\Sigma$ has a constant $k:1\to S$}$
\State $\mbox{closed}\gets \mbox{an array of size $n_s$ with $\mbox{closed[i]}=\mbox{false}$}$
\State $\mbox{count}\gets \mbox{an array of size $n_f$ with $\mbox{count[i]}=\mbox{the arity of $f_i$}$}$
\While{queue is not empty}
   \State $S\gets \mbox{pop}(\mbox{queue})$
   \If{$S=A$}       
      \State \Return true
   \EndIf
   \If{$\mbox{closed[S]}=\mbox{false}$}
      \State $\mbox{closed[S]}\gets \mbox{true}$
      \For{$i\gets 0$ to  $n_f-1$}
         \If{$S$ occurs in the type of $f_i$}
            \State $\mbox{count[i]}\gets \mbox{count[i]}-1$
            \If{$\mbox{count[i]}=0$} push $S$ onto queue
            \EndIf
         \EndIf
      \EndFor
   \EndIf
\EndWhile
\State \Return false
\EndProcedure
\end{algorithmic}
\end{algorithm}

Algorithm~[\ref{closed-sort}] is similar to Algorithm~[\ref{pl-fc}]\;:
identify sorts in the former with proposition symbols in the latter,
constants with axioms $\sequent{\top}{1}{R}$ and
function symbols with sequents.

\section{Forward Chaining for First-Order Horn Theories}\label{sec:fo-fc}

First-order inference involves 
 applying sequent rules for the relevant fragment of logic and discovering substitutions 
that allow the inference to proceed.
Before applying cut to derive of $\sequent{\top}{x_3}{C(x_3)}$ from axioms $\sequent{\top}{x_1}{A(x_1)}$ and
$\sequent{A(x_2)}{x_2}{B(x_2)}$, for example, we  apply $[x_3/x_1]$ and $[x_3/x_2]$ to the axioms.
We may also derive $\sequent{\top}{x_3, w}{C(x_3)}$ by applying the weakening substitution
$[x_3, w / x_3, w]$ regardless of the sort of $w$.
If, however, the first axiom were $\sequent{\top}{x_1, y}{A(x_1)}$, it is not clear that we can
answer the original query  $\sequent{\top}{x_3}{C(x_3)}$ since $y$ must be eliminated.

Algorithm~[\ref{fo-fc}] performs inference in first-order Horn theories.
It uses unification to discover substitutions and, if necessary, 
attempts to use the methods of Section~[\ref{closed-sorts}] to eliminate variables.

Consider a theory $\mathbb{T}$ with axioms (1)~$\sequent{A(x_1)}{x_1,\,y}{B(x_1)}$,
(2)~$\sequent{(B(x_2)\wedge C(x_2))}{x_2, w_1}{D(x_2, w_1)}$, and 
(3)~$\sequent{\top}{x_3}{A(x_3)}$.  We seek to derive the 
sequent $\sequent{C(x_4)}{x_4, w_2,\,z}{D(x_4, w_2)}$ where $w_i:W$, $x_i:X$, $y:Y$ and $z:Z$.
In applying  Algorithm~[\ref{fo-fc}], the initial queue has $x_3.\,A(x_3)$ 
and $x_4, w_2, z.\,C(x_4)$. 
The first pass through the while loop
unifies $[x_1, y.\,A(x_1)]$ and $[x_3.\,A(x_3)]$ by discovering $\theta_A=[y, x_3/ y, x_1]$.
This adds $y, x_3.\,B(x_3)$ to the queue.  The second pass unifies
\[[x_2, w_1.\,B(x_2),\; x_2, w_1.\,C(x_2)]\hspace{10pt}\mbox{and}\hspace{10pt}
  [y, x_3.\, B(x_3),\; x_4, w_2, z.\,C(x_4)]\]
by discovering $\theta_1=[w_1, y, x_3/ w_1, y, x_2]$ then $\theta_2=[w_1, y, z, w_2, x_4/w_1, y, z, w_2, x_3]$.
These compose to give the substitution $\theta_B=[w_1, y, x_4, z, w_2, x_4/w_1, y, x_2, z, w_2, x_3]$.
This adds $w_1, y, x_4, z, w_2.\,D(x_4, w_1)$ to the queue.  
Since line~[\ref{goalq-fo-fc}] discovers that this unifies with the goal $x_4, w_2,z.\,D(x_4, w_2)$, 
we seek to reconcile the formula-in-context with the goal.  $w_1$ and $w_2$ have
the same sort so the substitution $[w_2/w_1]$ leaves only $y$ as a variable in the derived context
that does not occur in the goal.
We then apply Algorithm~[\ref{closed-sort}] to determine if $Y$ is a closed sort in which 
case we could eliminate $y$ using a substitution $[k/y]$ with $k$ a constant.


\begin{algorithm}[h] 
\caption{Determine if a Horn sequent $\sigma=\sequent{(R_1\wedge\cdots \wedge R_k)}{\vec{y}}{S}$ 
    in normal form is derivable in a propositional Horn theory 
    $\mathbb{T}$ in normal form with sequents
    $\sequent{(P_1\,^i\wedge\dots\wedge P\,_{n_i}^i)}{\vec{x}\,^i}{Q^i}$ for $1\leq i\leq m$.\label{fo-fc}}
\begin{algorithmic}[1]
\Procedure{forward-chaining\,}{$\mathbb{T}$, $\sigma$}
\State $\mbox{queue}\gets$ a queue containing $\vec{y}.R_1$, $\dots$, $\vec{y}.R_k$ and 
        all $\vec{x}^{\;i}.\,Q^i$
    for which $(P_1^i\wedge \cdots \wedge P_{n_i}^i)=\top$
\State $\mbox{new}\gets \mbox{true}$
\While{$\mbox{new} = \mbox{true}$}
   \State $\mbox{new}\gets\mbox{false}$
   \For{$i\gets 1$ to  $m$}
      \For{each list $\vec{u}_1.\varphi_1$, $\dots$, $\vec{u}_{n_i}.\varphi_{n_i}$ from the queue
         \State $\theta\gets\Call{unify}{[\vec{u}_1.\varphi_1, \dots, \vec{u}_{n_i}.\varphi_{n_i}],
                 \,[\vec{x}^{\,i}.P_1^{\,i},\dots, \vec{x}^{\,i}.P_{n_i}^{\,i} ]}$}
         \If{$\theta\not=\mbox{Null}$}
            \State $\vec{v}.\,Q'\gets (\vec{x}^{\,i}.Q^i)\theta$
            \For{$\vec{u}.\varphi$ in the queue}
               \State $\gamma\gets \Call{unify}{[\vec{u}.\varphi],\,[\vec{v}.Q']}$
               \If{$\gamma\not=\mbox{Null}$}
                  \State add $\vec{v}.Q'$ to the queue
                  \State $\mbox{new}\gets\mbox{true}$
               \EndIf
             \EndFor
         \If{$\mbox{new}=\mbox{true}$}
            \State $\delta\gets \Call{unify}{[\vec{y}.S], [\vec{v}.Q']}$\label{goalq-fo-fc}
               \If{$\delta\not=\mbox{null}$}
                  \State $\vec{z}.S = (\vec{v}.Q')\delta$
                  \State $\mbox{reconcilable}\gets\mbox{true}$
                  \For{$z:Z$ with $z\in \vec{z}$ and $z\not\in\vec{y}$}
                      \If{$Z$ is not a closed sort and $\vec{y}$ has no $y:Z$}
                         \State $\mbox{reconcilable}\gets\mbox{false}$
                      \EndIf
                  \EndFor
                  \If{$\mbox{reconcilable}=\mbox{true}$} 
                    \State \Return true 
                  \EndIf
               \EndIf
         \EndIf
         \EndIf
       \EndFor
    \EndFor
\EndWhile
\State \Return false
\EndProcedure
\end{algorithmic}
\end{algorithm}
The example discussed above generates the following derivation.

\begin{tabular}{llll}
1 & $\sequent{C(x_4)}{x_4,\, w_1,\, z}{C(x_4)}$ & Id\\
2 & $\sequent{A(x_3)}{x_3,\, y}{B(x_3)}$      & Apply $\theta_A$ to Axiom 1\\
3 & $\sequent{\top}{x_3,\, y}{B(x_3)}$        & $\top$\\
4 & $\sequent{C(x_4)}{y,\, w_1,\, x_4,\, w_2,\, z}{C(x_4)}$ & Apply $\theta_B$ to 1\\
5 & $\sequent{(B(x_4)\wedge C(x_4))}{y,\, w_1,\, x_4,\, w_2,\, z}{D(x_4, w_1)}$ & Apply $\theta_B$ to Axiom 2\\
6 & $\sequent{C(x_4)}{y,\, w_1,\, x_4,\, w_2,\, z}{\top}$ & $\top$\\
7 & $\sequent{\top}{y,\, w_1,\, x_4,\, w_2,\, z}{B(x_4)}$ & Apply $\theta_B$ to 3\\
8 & $\sequent{C(x_4)}{y,\, w_1,\, x_4,\, w_2,\, z}{B(x_4)}$ & Cut 6, 7\\
9 & $\sequent{C(x_4)}{y,\, w_1,\, x_4,\, w_2,\, z}{(B(x_4)\wedge C(x_4))}$ & $\wedge$ 8, 4\\
10 & $\sequent{C(x_4)}{y,\, w_1,\, x_4,\, w_2,\, z}{D(x_4, w_1)}$ & Cut 9, 5\\
11 & $\sequent{C(x_4)}{y,\, w_2,\, x_4,\, z}{D(x_4, w_2)}$ & 10 $[w_2 / w_1]$\\[1pt]
\end{tabular}

If $y\!:\!Y$, $Y$ is a closed sort, and $k\!:\!Y$ is constant term, then the substitution $[k/y\,]$ 
yields the  goal~sequent.

In our AI courses we have used the Prover9 
\cite{Prover9} and Vampire \cite{Vampire} classical resolution-based theorem provers.
An implementation of the category theoretic algorithms discussed in this paper is ongoing.
\vfill
\eject

\nocite{*}
\bibliographystyle{plain}
\bibliography{act1}

\vfill
\eject

\section{Appendix}\label{sec:appendix}

This appendix includes (1) details of algorithms that are either called by 
algorithms in the body of the paper or that have been adapted, (2)~derivations 
using the sequent calculus of~\cite{Elephant}, and (3)~proofs of semantic
properties related to substitution that are referenced in the paper.







\subsection{Algorithms}

A C programming language implementation of the algorithms discussed in this paper will
be available at~\cite{sourcecode}. 
In our AI courses we have used the Prover9 
\cite{Prover9} (written in C) 
and Vampire \cite{Vampire} (written in C++) classical resolution-based theorem provers.

\subsubsection{Atomic Subformulae}

Algorithm~[\ref{atomic-subformulae}] returns a list of the atomic 
subformulae of a formula $\varphi$. Assume we have concatenation 
operation $+$ on lists.

\begin{algorithm}[h] 
\caption{Get the atomic subformulae of a formula $\varphi$.}\label{atomic-subformulae}
\begin{algorithmic}[1]
\Procedure{atomic-subformulae}{$\varphi$}
\If{$\varphi=\top$ or $\bot$ or $(t_1=t_2)$ or $R(t_1, \dots, t_n)$}
   \State \Return a list with one element $\varphi$
\ElsIf{$\varphi = \varphi_0*\varphi_1$ with $*=\wedge$ or $\vee$ or $\Rightarrow$}
   \State \Return \Call{atomic-subformulae}{$\varphi_0$} $+$ \Call{atomic-subformulae}{$\varphi_1$}
\ElsIf{$\varphi=\neg\varphi_0$}
   \State \Return \Call{atomic-subformulae}{$\varphi_0$}
\Else{$\varphi=(Q\,x)(\varphi_0)$ where $Q=\exists$ or $\forall$}
   \State \Return \Call{atomic-subformulae}{$\varphi_0$}
\EndIf
\EndProcedure
\end{algorithmic}
\end{algorithm}

This algorithm, together with an equality function on formula, 
is  used to implement a \textsc{Unique-Atomic-Subformulae} function.

\subsubsection{Substituting a Term for a Variable in a Term}\label{sub-s-for-x-in-t}

Substituting a term $s$ for a variable $x$ in a term $t$ involves first checking the class of the $t$.
If $t$ is a variable then either $t=x$ or $t$ is a different variable.  In the former case, substitution
returns $s$.  In the latter case it returns $t$.  If $t$ is a function term $f(u_1,\, \dots,\, u_n)$, we
 construct a new function term and  substitute $s$ for $x$ in each $u_i$.
\vspace{-3pt}

\begin{algorithm}
\caption{Substitute a term $s:A$ for a variable $x:A$ in a term $t$.}\label{term-sub}
\begin{algorithmic}[1]
\Procedure{term-substitution}{$t$, $s$, $x$}
\If{$\mbox{sort}(s)\not=\mbox{sort}(x)$}
   \State \Return Null
\EndIf
\If{$t=x$} 
    \Return $s$
\ElsIf{$t=$ a variable $y$ and $y\not=x$}
    \Return $t$
\Else{\ $t=f(u_1, \dots, u_n)$}
   \State \Return $f(\Call{term-substitution}{u_1, s,\, x}, \dots, \Call{term-substitution}{u_n, s,\, x})$
\EndIf
\EndProcedure
\end{algorithmic}
\end{algorithm}\vspace{-15pt}
\vfill
\eject

\subsubsection{Simultaneous Substitutions into a Term}\label{simultaneous-sub-t}

To implement simultaneous substitution of the terms $\vec{s}$ for 
variables $\vec{y}$ 
we must ensure that
$s_j$ is substituted only for occurrences of $y_j$ in $t$ and not for
occurrences that might arise after substituting $s_i$ for $y_i$ with $i\not=j$.
That is, the output of the algorithm must be the same as if the 
individual substitutions were made simultaneously.

\begin{algorithm}
\caption{Apply a substitution $\theta=[s_1, \dots, s_n / y_1, \dots , y_n]$ 
     to a term $t$}\label{simsub}
\begin{algorithmic}[1]
\Procedure{term-substitution}{$t$, $\theta$}
\If{$t=y_i$ for some variable $y_i$ of $\theta$}
   \State \Return $s_i$
\ElsIf{$t=y$ for some variable $y\not= y_i$}
    \State \Return $t$
\ElsIf{$t=f(u_1, \dots u_n)$}
    \State $t_0=t$
    \For{$i\gets 1$ to  $n$}
       \State Assume $y_i:Y$
       \State $y_i':Y\gets$ distinct from $y_1',\dots, y_{i-1}'$,
                    $y_{i+1},\,\dots,\,y_n$ and all variables in $\vec{s}$\label{simsub-yp}
       \State $t_i\gets t_{i-1}[y_i'\,/\,y_i]$
    \EndFor
    \State $t_0' \gets t_n$
    \For{$i\gets 1$ to $n$}\label{simsub-loop2a}
       \State $t_i'\gets t_{i-1}'[s_i\,/\,y_i']$ \label{simsub-loop2b}
    \EndFor
    \State \Return $t_n'$
\EndIf
\EndProcedure
\end{algorithmic}
\end{algorithm}

Line~[\ref{simsub-yp}] of Algorithm~[\ref{simsub}] ensures that the 
$y_i'$ variables are distinct from each other and from all variables of $\vec{s}$.
The sequential substitutions $[s_i\,/\,y_i'\,]$ 
applied in  lines~[\ref{simsub-loop2a}]--[\ref{simsub-loop2b}], 
therefore, result in the same output as
would be achieved by a simultaneous substitution.

\subsubsection{Substitution into a Term-in-Context}

\begin{algorithm}
\caption{Apply a substitution $\theta=[s_1, \dots, s_n / y_1, \dots , y_n]$ 
     to a term-in-context $\vec{z}.t$}\label{alg:sub-term-in-context}
\begin{algorithmic}[1]
\Procedure{term-in-context-substitution}{$\vec{z}.t$, $\theta$}
\State $\vec{z}\,'=$ all variables of $\vec{z}$ that are not in $\vec{y}$
\State $\theta^{\vec{z}} = [\vec{s}, \vec{z}\,'/\vec{y}, \vec{z}\,']$
\State $\vec{x}=$ the canonical context for $\theta^{\vec{z}}$
\State $t_1=$\Call{term-substitution}{$t$, $\theta^{\vec{z}}$}
\State \Return $\vec{x}.\,t_1$
\EndProcedure
\end{algorithmic}
\end{algorithm}
\vfill
\eject

\subsubsection{Substituting a Term for a Variable in a Formula}\label{sub-s-for-x-in-phi}


When substituting terms for variables in a formula, one must ensure that variables of
the terms do not fall within the scope of a quantifier. We can avoid this
by changing the quantified variable to one that does not occur in the terms to be substituted
or elsewhere in the formula.  Here are three examples.\vspace{-8pt}
\begin{align*}
(\exists z)\varphi(x, z)\; [f(y)/z\,] &= (\exists z')\varphi(x, z') \\
(\exists z)\varphi(x, z)\; [f(y)/x\,] &= (\exists z')\varphi(f(y), z')\\
(\exists z)\varphi(x, z)\; [f(z)/x\,] &= (\exists z')\varphi(f(z), z')\vspace{-12pt}
\end{align*}

Assume we have a VariableFactory data structure that can be initialized for a specified sort $A$
and which has a method next() to successively generate distinct new variables of sort $A$.

\begin{algorithm}[h]
\caption{Substitute a term $s:A$ for a variable $x:A$ in a formula $\varphi$.}\label{formula-sub}
\begin{algorithmic}[1]
\Procedure{formula-substitution}{$\varphi$, $s$, $x$}
\If{$\varphi = \top$ or $\bot$}
    \Return $\varphi$
\ElsIf{$\varphi = R(t_1, \dots, t_n)$}
   \State \Return $R(\mbox{term-substitution}(t_1, s, x), \dots, \mbox{term-substitution}(t_n, s, x))$
\ElsIf{$\varphi = (t_0 = t_1)$}
   \State \Return $(\mbox{term-substitution}(t_0, s, x) = \mbox{term-substitution}(t_0, s, x))$
\ElsIf{$\varphi = (\varphi_0 * \varphi_1)$ with $*=\wedge$, $\vee$ or $\Rightarrow$}
 \State \Return $(\mbox{formula-substitution}(\varphi_0, s, x) *\mbox{formula-substitution}(\varphi_1, s, x))$
\ElsIf{$\varphi = (\neg \varphi_0)$}
   \State \Return $(\neg \mbox{formula-substitution}(\varphi_0, s, x))$
\ElsIf{$\varphi = (Qz).\varphi_0$ with $Q=\exists$ or $\forall$}
   \State $\mbox{vf} \gets $ new VariableFactory for sort $Z$ where $z:Z$
  \While{$z'=\mbox{vf}.\mbox{next}()$  occurs in $s$ or $\phi_0$}
    \State continue
  \EndWhile
  \State $\phi_0'\gets \mbox{formula-substitution}(\phi_0, z', z)$
  \State $\phi_0''\gets \mbox{formula-substitution}(\phi_0', s, x)$
  \State \Return $(Q.z')\phi_0''$
\EndIf
\EndProcedure
\end{algorithmic}
\end{algorithm}\vspace{-15pt}

\subsubsection{Simultaneous Substitutions into a Formula} \label{simultaneous-sub-phi}

The formula substitution algorithms  rely on a VariableFactory data structure that
generates new variables of a specified type and a method for testing equality of variables.
Given a substitution $[\vec{s}/\vec{x}]$, once we have ensured that no $x_i\in\vec{x}$ 
occurs in $\vec{s}$ of elsewhere in $\vec{x}$, then we can apply the substitutions sequentially.

\begin{algorithm}[h]
\caption{Apply a  substitution $\theta=[s_1, \dots, s_n / x_1, \dots , x_n]$ 
     to a formula $\varphi$}\label{simsubphi}
\begin{algorithmic}[1]
\Procedure{formula-substitution}{$\varphi$, $\theta$}
\For{$x\;\mbox{in}\;\vec{x}$}
   \State Find a variable $x_i'$ of the same sort as $x$ but not occuring in $\vec{s}$ or $\varphi$
     \State $\Call{formula-substitution}{\varphi,\, x_i',\, x_i}$
     \State Replace $[s_i/x_i]$ by $[s_i/x_i'\,]$ in $[\vec{s}/\vec{x}\,]$.
   \EndFor
\For{$x\;\mbox{in}\;\vec{x}$}
    \State $\Call{formula-substitution}{\varphi,\,s_i,\,x_i'}$
\EndFor   
\EndProcedure
\end{algorithmic}
\end{algorithm}
\vfill
\eject

\subsubsection{Substitution into a Formula-in-Context}

We need only extend the substitution then call Algorithm~[\ref{simsubphi}].

\begin{algorithm}[h]
\caption{Apply a substitution $\theta=[s_1, \dots, s_n / y_1, \dots , y_n]$ 
     to a formula-in-context $\vec{z}.\varphi$}\label{alg:sub-formula-in-context}
\begin{algorithmic}[1]
\Procedure{formula-in-context-substitution}{$\vec{z}.\varphi$, $\theta$} 
\State $\vec{z}\,'=$ all variables of $\vec{z}$ that are not in $\vec{y}$
\State $\theta^{\vec{z}} = [\vec{s}, \vec{z}\,'/\vec{y}, \vec{z}\,']$
\State $\vec{x}=$ the canonical context for $\theta^{\vec{z}}$
\State \Return $\vec{x}.\Call{formula-substitution}{\varphi, \theta^{\vec{z}}}$
\EndProcedure
\end{algorithmic}
\end{algorithm}


\subsubsection{Normal Form for a Horn Theory}\label{sec:horn-theory-normal-form}

We assume a  data structure
such as a hash set that efficiently implements insertions~\cite{CLRS}.

\begin{algorithm}
\caption{Convert a Horn sequent $\sigma =\sequent{\varphi}{\vec{x}}{\psi}$ to normal form.}\label{horn-normal}
\begin{algorithmic}[1]
\Procedure{Horn-normal-form}{$\sigma$}
\State $\alpha[n] \gets$ \Call{unique-atomic-subformulae}{$\psi$}.  
         See Algorithm~[\ref{atomic-subformulae}] in the appendix.
\State $\Gamma[n] \gets$ an uninitialized array of sequents
\For{$i\gets 0$ to  $n-1$}
   \State $\Gamma[i] = \sequent{\varphi}{\vec{x}}{\alpha[i]}$
\EndFor
\State \Return $\Gamma$
\EndProcedure
\end{algorithmic}
\end{algorithm}

Distinct axioms of $\mathbb{T}$ may generate identical sequents in their normal forms.
We again rely on a data structure to ignore redundant insertions.
\begin{algorithm}[h]
\caption{Convert a Horn theory $\mathbb{T}$ to normal form.}\label{horn-theory-normal}
\begin{algorithmic}[1]
\Procedure{Horn-theory-normal-form}{$\mathbb{T}$}
\State $\mbox{axioms}\gets\{\,\}$
\For{$\sigma\in\mathbb{T}$}
   \State $\mbox{axioms}\gets\mbox{axioms}\cup \Call{Horn-normal-form}{\sigma}$
\EndFor
\State\Return axioms
\EndProcedure
\end{algorithmic}
\end{algorithm}\vspace{-15pt}

\vfill
\eject

\subsubsection{Unification of Lists of Terms}\label{unify1}

A unification of lists $[\alpha_1, \dots, \alpha_m]$ and $[\beta_1, \dots, \beta_m]$ of 
terms is a substitution $\theta$ for which $\alpha_i\,\theta = \beta_i\,\theta$ for $1\leq i\leq m$.
Algorithm~[\ref{unification}] adapts Definition 7.2.9 and Theorem 7.2.11 of~\cite{BPT}
to support multi-sorted signatures.

\begin{algorithm}
\caption{Find a unification $\theta$ of two lists of terms.}\label{unification}
\begin{algorithmic}[1]
\Procedure{unify-terms}{$[\alpha_1, \dots, \alpha_m],\, [\beta_1, \dots, \beta_n]$, $\theta=[]$}
\If{$m\not=n$} \Return Null
\ElsIf{$m = 0$} \Return $\theta$
\ElsIf{$\mbox{sort}(\alpha_1)\not=\mbox{sort}(\beta_1)$} \Return Null
\ElsIf{$\alpha = x:X$ is a variable}
  \If{$\beta_1=y$ is a variable}
     \If{$\alpha_1=\beta_1$} 
           \Return \Call{unifty-terms}{$[\alpha_2, \dots, ],\,[\beta_2, \dots ],\, \theta$}
     \Else{} \Return \Call{unify-terms}{$[\alpha_2[\beta_1/\alpha_1], \dots ],
           \, [\beta_2[\beta_1/\alpha_1], \dots ],\; \theta [\beta_1/\alpha_1]$}
     \EndIf
  \Else {\ $\beta_1 = g(t_1, \dots, t_\ell)$}
     \If{$\alpha_1$ occurs in $\beta_1$} \Return Null
     \Else{} \Return \Call{unify-terms}{$[\alpha_2[\beta_1/\alpha_1], \dots ],
           \, [\beta_2[\beta_1/\alpha_1], \dots ],\; \theta [\beta_1/\alpha_1]$}
     \EndIf
  \EndIf
\Else {\ $\alpha_1 = f(s_1, \dots, s_k)$}
   \If{$\beta_1$ is a variable} \Return \Call{unify-terms}{$[\beta_1, \alpha_2, \dots],\,
                  [\alpha_1, \beta_2, \dots],\,\theta$}
   \Else {\ $\beta_1 = g(t_1, \dots, t_\ell)$}
      \If{$f\not= g$} \Return Null
      \Else{}  \Return \Call{unify-terms}{$[s_1, \dots, s_k, \alpha_2, \dots],\,
                                       [t_1, \dots, t_\ell, \beta_2, \dots],\,\theta$}
      \EndIf
    \EndIf
\EndIf
\EndProcedure
\end{algorithmic}
\end{algorithm}

Here is an example in which the procedure finds a unification.

  \begin{tabular}{llll}
  1 & $\mbox{\function{unify-terms}}([g(x),f(y)],\; [g(f(y)), f(g(z))],\; [\,])$\span\span\\[2pt]
  2 & $\mbox{\function{unify-terms}}([x, f(y)],\; [f(y), f(g(z))],\; [\,])$\span\span\\[2pt]
  3 & $\mbox{\function{unify-terms}}([f(y)],\; [f(g(z))],\; [f(y)/x])$\span\span\\[2pt]
  4 & $\mbox{\function{unify-terms}}([y],\; [g(z)],\; [f(y)/x])$\span\span\\[2pt]
  5 & $\mbox{\function{unify-terms}}([\,],\; [\, ],\; [f(y)/x]\,[g(z)/y])$\span\span\\[3pt]
  Check:\span\span\span\\
  & $g(x)[f(y)/x]\,[g(z)/y])$    & $= g(f(y)) [g(z)/y]$ & $= g(f(g(z))$\\[2pt]
  & $g(f(y))[f(y)/x]\,[g(z)/y])$ & $= g(f(y)) [g(z)/y]$ & $= g(f(g(z))$\\[4pt]
  & $f(y)[f(y)/x]\,[g(z)/y])$    & $= f(y) [g(z)/y]$    & $= f(g(z))$\\[2pt]
  & $f(g(z))[f(y)/x]\,[g(z)/y])$ & $= f(g(z)) [g(z)/y]$ & $= f(g(z))$\\[3pt]
  \end{tabular}

Here is an example in which no unification can be found.

  \begin{tabular}{llll}
  1 & $\mbox{\function{unify-terms}}([f(x, g(x))],\, [f(y,y)], \,[\,])$\span\span\\[2pt]
  2 & $\mbox{\function{unify-terms}}([x, g(x)],\,[y, y],\, [\,])$\span\span\\[2pt]
  3 & $\mbox{\function{unify-terms}}([g(y)],\,[y],\,[y/x])$\span\span\\[2pt]
  4 & $\mbox{\function{unify-terms}}([y],\,[g(y)],\,[y/x])$\span\span\\[2pt]
  5 & Null
  \end{tabular}
\vfill
\eject

\subsubsection{Unification of Lists of Formulae}\label{unify2}
\begin{algorithm}[h]
\caption{Find a unification $\theta$ of two lists of formulae.}\label{unificationphi}
\begin{algorithmic}[1]
\Procedure{unify-formulae}{$[\alpha_1, \dots, \alpha_m],\,[\beta_1,\dots, \beta_n], \theta=[])$}
\If{$m\not= n$} \Return Null
\ElsIf{$m=0$} \Return $\theta$
\ElsIf{$\alpha_1=(s_1 = s_2)$ and $\beta_1=(t_1 = t_2)$} 
  \State $\theta' = \mbox{\Call{unify-terms}{$[s_1, s_2],\, [t_1, t_2],\,\theta$}}$
  \State \Return \Call{unify-formulae}{$[\alpha_2\,\theta', \dots],\,[\beta_2\,\theta',\dots],\,\theta\,\theta'$}
\ElsIf {$\alpha_1=Q(s_1, \dots, s_k)$ and $\beta_1=R(t_1, \dots, t_\ell)$}
   \If {$Q\not=R$} \Return Null
   \Else
      \State $\theta'=\mbox{\Call{unify-terms}{$[s_1, \dots, s_k],\,[t_1, \dots, t_k], \theta$}}$
      \State \Return \Call{unify-formulae}{$[\alpha_2\theta', \dots, ],\,[\beta_2\,\theta',\dots], \theta\,\theta'$}
    \EndIf
\ElsIf {$\alpha_1 = \varphi*\varphi'$ and $\beta_1=\psi * \psi'$ with $*=\wedge$, $\vee$ or $\Rightarrow$}
   \State \Return \Call{unify-formulae}{$[\varphi, \varphi', \alpha_2, \dots],\, 
                          [\psi,\psi', \beta_2,\dots],\, \theta$}
\ElsIf{$\alpha_1=\neg\varphi$ and $\beta_1=\neg\psi$}
   \State \Return \Call{unify-formulae}{$[\varphi, \alpha_2, \dots],\, 
                          [\psi, \beta_2,\dots],\, \theta$}
\ElsIf{$\alpha_1=((Q.x)\varphi)$ and $\beta_1=((Q.y)\psi)$ with $Q=\exists$ or $\forall$}
   \If{$\mbox{sort}(x)\not=\mbox{sort}(y)$} \Return Null
   \Else{} 
     \State Let $x_1$ and $x_2$ be variables of $\mbox{sort}(x)$ distinct from all variables in 
             $\alpha$ and $\beta$
     \If{$x\not=y$}  
            \State let $x_3$ be another such variable else let $x_3=??$
            \State \Return 
           \textsc{unify-formulae}$([\alpha_1[x_1/x][x_3/y], \alpha_2[x_2/x, x_3/y], \dots],\;$
           \State\hphantom{return \ \,\textsc{unify-formulae}}  $[\beta_1[x_1/y][x_2/x], \beta_2[x_2/x, x_3/y], \dots],\;
                                                 \theta[x_2, x_3 / x, y]$
     \EndIf
   \EndIf
\EndIf
\State \Return Null
\EndProcedure
\end{algorithmic}
\end{algorithm}

  \begin{tabular}{lllll}
  1 & $\function{Unify-Formulae}([((\exists x). R(x, y)),\, P(x)],\; 
        [((\exists y). R(y, z)),\, P(t)],\; [\,])$\span\span\span\\
  2 & $\function{Unify-Formulae}([R(x_1, x_3),\, P(x_2)],\; [R(x_1, z),\, P(t)], \; 
          [x_2, x_3 / x, y])$\span\span\span\\
  3 & $\theta'=\function{Unify-Terms}([x_1, x_3],\; [x_1, z],\; [x_2, x_3 / x, y])$\span\span\span\\
  4 & $\theta'=\function{Unify-Terms}([x_3],\; [z],\; [x_2, x_3 / x, y])$\span\span\span\\
  5 & $\theta'= [x_2, x_3 / x, y]\, [z/x_3]$\span\span\span\\
  6 & $\function{Unify-Formulae}([P(x_2)],\; [P(t)], \; 
          [x_2, x_3 / x, y]\, [z/x_3])$\span\span\span\\
  7 & $\function{Unify-Formulae}([\,],\; [\,], \; 
          [x_2, x_3 / x, y]\, [z/x_3]\, [t/ x_2])$\span\span\span\\
  8 & $ [x_2, x_3 / x, y]\, [z/x_3]\, [t/ x_2]$\span\span\span\\[3pt]
Check:  & $((\exists x).R(x, y)) [x_2, x_3 / x, y]\, [z/x_3]\, [t/ x_2]$ 
      & $=((\exists x).R(x, x_3))[z/x_3]\, [t/ x_2]$ \\ 
      && $=((\exists x).R(x, z))[t/ x_2]
      =((\exists x).R(x, z))$ \\[2pt]
  & $((\exists y).R(y, z)) [x_2, x_3 / x, y]\, [z/x_3]\, [t/ x_2]$ 
      & $=((\exists y).R(y, x_2))[z/x_3]\, [t/ x_2]$ \\
      && $=((\exists y).R(y, z))[t/ x_2]
      =((\exists y).R(y, z))$\\[2pt]
  & $P(x)[x_2, x_3 / x, y]\, [z/x_3]\, [t/ x_2]$ 
      & $=P(x_2)[z/x_3]\, [t/ x_2]$ \\
      && $=P(x_2)[t/ x_2]
      =P(t)$\\[2pt]
  & $P(t)[x_2, x_3 / x, y]\, [z/x_3]\, [t/ x_2]$ 
      & $=P(t)[z/x_3]\, [t/ x_2]$ \\
      && $=P(t)[t/ x_2]
      =P(t)$\\
  \end{tabular}

\vfill
\eject

\subsection{Derivations}\label{derivations}

In this section we complete sequent calculus exercises set aside for
the reader in~\cite{Elephant}  and prove  results that arise
in developing algorithms for this paper.

\subsubsection{Equality}
Symmetry and transitivity of equality are derived inference rules.

\begin{theorem}
In Horn logic we can derive $\sequent{(x=y)}{x,y}{(y=x)}$.
\end{theorem} 

    \begin{tabular}{lll}
    1 & $\sequent{(x=y)}{x,y}{(x=y)}$ & ID\\[2pt]
    2 & $\sequent{(x=y)}{x,y}{\top}$  & $\top$\\[2pt]
    3 & $\sequent{\top}{x}{(x=x)}$  & Eq0\\[2pt]
    4 & $\sequent{\top}{x,y}{(x=x)}$  & Weakening, 3\\[2pt]

    5 & $\sequent{(x=y)}{x,y}{(x=x)}$ & Cut 2, 4\\
    6 & $\sequent{(x=y)}{x,y}{((x=y)\wedge (x=x))}$ & $\wedge$I 1, 5\\[2pt]
    7 & $\sequent{((x=y)\wedge(x=z))}{x,y,z}{(y=z)}$  & Eq1\\[2pt]
    8 & $\sequent{((x=y)\wedge(x=x))}{x,y}{(y=x)}$    & Sub $8[x/z]$\\[2pt]
    9 & $\sequent{(x=y)}{x,y}{(y=x)}$ & Cut 6, 8
    \end{tabular}\vspace{3pt}
\smallskip

\begin{theorem}
In Horn logic we can derive
$\sequent{((x=y)\wedge (y=z))}{x,y,z}{(x=z)}$.
\end{theorem}

    \begin{tabular}{lll}
    1 & $\sequent{((x=y)\wedge (y=z))}{x,y,z}{(x=y)}$ & $\wedge E0$\\[2pt]
    2 & $\sequent{((x=y)\wedge (y=z))}{x,y,z}{(y=z)}$ & $\wedge E1$\\[2pt]
    3 & $\sequent{((x=y)\wedge (y=z))}{x,y,z}{((y=z)\wedge (x=y))}$ & $\wedge$ 2, 1\\[2pt]
    4 & $\sequent{((y=z)\wedge (x=y))}{x,y,z}{(x=z)}$ & Eq1\\[2pt]
    5 & $\sequent{((x=y)\wedge (y=z))}{x,y,z}{(x=z)}$ & Cut 3, 4
    \end{tabular}

\subsubsection{Context Permutation}

\begin{theorem}
In atomic logic we can derive $\sequent{\varphi}{y,x}{\psi}$
from $\sequent{\varphi}{x,y}{\psi}$.
\end{theorem}

    \begin{tabular}{lll}
    1 & $\sequent{\varphi}{x,y}{\psi}$ & Hypothesis\\[2pt]
    2 & $\sequent{\varphi[x,y/x,y]}{x,y,z}{\psi[x,y/x,y]}$ & Sub1\\[2pt]
    3 & $\sequent{\varphi}{x,y,z}{\psi}$ & Restate 2\\[2pt]
    4 & $\sequent{\varphi[z,y,z/x,y,z]}{y,z}{\psi[z,y,z/x,y,z]}$ & Sub3\\[2pt]
    5 & $\sequent{\varphi}{y,z}{\psi}$ & Restate 4\\[2pt]
    6 & $\sequent{\varphi[x/z]}{y,x}{\psi[x/z]}$ & Sub3\\[2pt]
    7 & $\sequent{\varphi}{y,x}{\psi}$ & Restate 6
   \end{tabular}\\[3pt]
For example, $\sequent{R(x,y)}{x,y}{S(x,y)}$ and $\sequent{R(x,y)}{y,x}{S(x,y)}$
are provably equivalent.  It follows that we may permute the order of the
variables in the context of a sequent.
\vfill
\eject

\subsubsection{Commutative and Associative Laws}\label{assoc}

\begin{theorem}
In Horn logic $\sequentS{(\varphi\wedge\psi)}{\vec{x}}{(\psi\wedge \varphi)}$.
\end{theorem}

    \begin{tabular}{lll}
    1 & $\sequent{(\varphi\wedge\psi)}{\vec{x}}{\varphi}$ & $\wedge$E0\\[2pt]
    2 & $\sequent{(\varphi\wedge\psi)}{\vec{x}}{\psi}$ & $\wedge$E1\\[2pt]
    3 & $\sequent{(\varphi\wedge\psi)}{\vec{x}}{(\psi\wedge \varphi)}$ & $\wedge$ 1,2
    \end{tabular}
\smallskip

\begin{theorem}
In Horn logic 
$\sequentS{(\varphi\wedge(\psi\wedge\chi))}{\vec{x}}{((\varphi\wedge \psi)\wedge \chi)}$.
\end{theorem}

    \begin{tabular}{lll}
    1 & $\sequent{(\varphi\wedge(\psi\wedge\chi))}{\vec{x}}{\varphi}$ & $\wedge$E0\\[2pt]
    2 & $\sequent{(\varphi\wedge(\psi\wedge\chi))}{\vec{x}}{(\psi\wedge\chi)}$ & $\wedge$E1\\[2pt]
    3 & $\sequent{(\psi\wedge\chi)}{\vec{x}}{\psi}$ & $\wedge$E0\\[2pt]
    4 & $\sequent{(\varphi\wedge(\psi\wedge\chi))}{\vec{x}}{\psi}$ & Cut 2, 3\\[2pt]
    5 & $\sequent{\varphi\wedge (\psi\wedge\chi))}{\vec{x}}{(\varphi\wedge \psi)}$ & $\wedge$ 1, 4\\[5pt]
    6 & $\sequent{(\psi\wedge\chi)}{\vec{x}}{\chi}$ & $\wedge$E1\\[2pt]
    7 & $\sequent{\varphi\wedge (\psi\wedge\chi))}{\vec{x}}{\chi}$ & Cut 2, 6\\[2pt]
    8 & $\sequent{(\varphi\wedge(\psi\wedge\chi))}{\vec{x}}{((\varphi\wedge \psi)\wedge \chi)}$
            & $\wedge$ 5, 8
    \end{tabular}
\smallskip

\begin{theorem}
In coherent logic $\sequentS{(\varphi\vee\psi)}{\vec{x}}{(\psi\vee \varphi)}$.
\end{theorem}

    \begin{tabular}{lll}
    1 & $\sequent{\varphi}{\vec{x}}{(\varphi\vee\psi)}$ & $\wedge$I0\\[2pt]
    2 & $\sequent{\psi}{\vec{x}}{(\varphi\vee\psi)}$ & $\wedge$I0\\[2pt]
    3 & $\sequent{(\varphi\vee\psi)}{\vec{x}}{(\psi\vee \varphi)}$ & $\wedge$ 2, 1
    \end{tabular}
\smallskip

\begin{theorem}
In coherent logic 
$\sequentS{(\varphi\vee(\psi\vee\chi))}{\vec{x}}{((\varphi\vee \psi)\vee \chi)}$.
\end{theorem}

    \begin{tabular}{lll}
    1 & $\sequent{\varphi}{\vec{x}}{(\varphi\vee \psi)}$ & $\vee$I0\\[2pt]
    2 & $\sequent{(\varphi\vee\psi)}{\vec{x}}{((\varphi\vee \psi)\vee \chi)}$ & $\vee$I0\\[2pt]
    2 & $\sequent{\varphi}{\vec{x}}{((\varphi\vee \psi)\vee \chi)}$ & Cut 1, 2\\[5pt]
    3 & $\sequent{\psi}{\vec{x}}{(\varphi\vee\psi)}$ & $\vee$I1\\[2pt]   
    4 & $\sequent{(\varphi\vee\psi)}{\vec{x}}{((\varphi\vee\psi)\vee \chi)}$ & $\vee$I0\\[2pt]   
    5 & $\sequent{\psi}{\vec{x}}{((\varphi\vee\psi)\vee \chi)}$ & Cut 3, 4\\[2pt]   
    6 & $\sequent{\chi}{\vec{x}}{((\varphi\vee\psi)\vee \chi)}$ & $\vee$I1\\[2pt]   
    7 & $\sequent{(\psi\vee\chi)}{\vec{x}}{((\varphi\vee\psi)\vee \chi)}$ & $\vee$ 5, 6\\[2pt]   
    8 & $\sequent{(\varphi\vee(\psi\vee\chi))}{\vec{x}}{((\varphi\vee \psi)\vee \chi)}$
           & $\vee$ 2, 7
    \end{tabular}

\subsubsection{Rules with $\top$ and $\bot$}\label{telim}

\begin{theorem}\label{thm:topwedge}
In Horn logic, $\sequentS{(\top\wedge \varphi)}{\vec{x}}{\varphi}$.
\end{theorem}

    \begin{tabular}{lllclll}
    1 & $\sequent{(\top\wedge\varphi)}{\vec{x}}{\varphi}$ & $\wedge$E0 
             &\hspace{.5in} &
              1 & $\sequent{\varphi}{\vec{x}}{\varphi}$ & ID\\[2pt]
      & & & & 2 & $\sequent{\varphi}{\vec{x}}{\top}$    & $\top$\\[2pt]
      & & & & 3 & $\sequent{\varphi}{\vec{x}}{(\top\wedge\varphi)}$ & $\wedge$I 1, 2
   \end{tabular}

\vfill
\eject

\begin{theorem}\label{thm:topwedgecor}
In Horn logic, 
$\displaystyle\frac{\underline{\sequent{(\top\wedge \varphi)}{\vec{x}}{\psi}}}{
\sequent{\varphi}{\vec{x}}{\psi}}$
\end{theorem}

    \begin{tabular}{lllclll}
     1 & $\sequent{(\top\wedge \varphi)}{\vec{x}}{\psi}$ & Hypothesis
          &\hspace{.5in} & 
           1 & $\sequent{\varphi}{\vec{x}}{\psi}$ & Hypothesis\\[2pt]
     2 & $\sequent{\varphi}{\vec{x}}{(\top\wedge\varphi)}$ & Theorem~[\ref{thm:topwedge}]
          & & 
           2 & $\sequent{(\top\wedge\varphi)}{\vec{x}}{\varphi}$ & Theorem~[\ref{thm:topwedge}]\\[2pt]
     3 & $\sequent{\varphi}{\vec{x}}{\psi}$ & Cut 2, 1
          & & 
           3 & $\sequent{(\top\wedge\varphi)}{\vec{x}}{\psi}$ & Cut 2, 1
    \end{tabular}

\begin{theorem}\label{thm:botvee}
In coherent logic, $\sequentS{(\bot\vee \varphi)}{\vec{x}}{\varphi}$.
\end{theorem}

    \begin{tabular}{lllclll}
    1 & $\sequent{\varphi}{\vec{x}}{(\bot\vee\varphi)}$ & $\vee$I0 
             &\hspace{.5in} &
              1 & $\sequent{\varphi}{\vec{x}}{\varphi}$ & ID\\
      & & & & 2 & $\sequent{\bot}{\vec{x}}{\varphi}$    & $\bot$\\
      & & & & 3 & $\sequent{(\bot\vee\varphi)}{\vec{x}}{\varphi}$ & $\vee$ 1, 2
    \end{tabular}

\begin{theorem}
In coherent logic, 
$\displaystyle\frac{\underline{\sequent{(\bot\vee \varphi)}{\vec{x}}{\psi}}}{
\sequent{\varphi}{\vec{x}}{\psi}}$
\end{theorem}

    \begin{tabular}{lllclll}
     1 & $\sequent{(\bot\vee \varphi)}{\vec{x}}{\psi}$ & Hypothesis
          &\hspace{.5in} & 
           1 & $\sequent{\varphi}{\vec{x}}{\psi}$ & Hypothesis\\[2pt]
     2 & $\sequent{\varphi}{\vec{x}}{(\bot\vee\varphi)}$ & Theorem~[\ref{thm:botvee}]
          & & 
           2 & $\sequent{(\bot\vee\varphi)}{\vec{x}}{\varphi}$ & Theorem~[\ref{thm:botvee}]\\[2pt]
     3 & $\sequent{\varphi}{\vec{x}}{\psi}$ & Cut 2,1
          & & 
           3 & $\sequent{(\bot\vee\varphi)}{\vec{x}}{\psi}$ & Cut 2,1
    \end{tabular}

\subsubsection{Frobenius Axiom}

\begin{theorem}\label{thm:Frobenius}
In intuitionistic logic 
we can derive the Frobenius Axiom. If $y$ does not occur in the context $\vec{x}$ and is not free in $\varphi$ 
  then:\label{Frobenius}
$\sequent{(\varphi\wedge(\exists y)\psi)}{\vec{x}}{%
    (\exists y)(\varphi\wedge\psi)}$.
\end{theorem}

    \begin{tabular}{lll}
     1 & $\sequent{((\exists y)(\varphi\wedge\psi))}{\vec{x}}{%
         ((\exists y)(\varphi\wedge\psi))}$ & ID\\[2pt]
     2 & $\sequent{(\varphi\wedge\psi)}{\vec{x}{y}}{%
         ((\exists y)(\varphi\wedge\psi))}$ & $\exists$ 1\\[2pt]
     3 & $\sequent{\psi}{\vec{x},y}{%
         (\varphi\Rightarrow((\exists y)(\varphi\wedge\psi)))}$
           & $\Rightarrow$ 2\\[2pt]
     4 & $\sequent{((\exists y)\psi)}{\vec{x}}{%
         (\varphi\Rightarrow((\exists y)(\varphi\wedge\psi)))}$
           & $\exists$ 3\\[2pt]
     5 & $\sequent{(\varphi\wedge (\exists y)\psi)}{\vec{x}}{%
         ((\exists y)(\varphi\wedge\psi))}$
           & $\Rightarrow$ 4\\[5pt]
    \end{tabular}

\begin{theorem}\label{thm:Frobeniusconverse}
In regular logic 
we can derive the converse of Frobenius.
If $y$ does not occur in the context $\vec{x}$ and is not free in $\varphi$ 
  then:\label{Frobenius-converse}
$\sequent{((\exists y)(\varphi\wedge\psi))}{\vec{x}}{%
       ((\varphi\wedge(\exists y)\psi))}$.
\end{theorem}

    \begin{tabular}{lll}
     1 & $\sequent{((\exists y)\psi)}{\vec{x}}{((\exists y)\psi)}$ & ID\\[2pt]
     2  & $\sequent{((\exists y)\psi)}{\vec{x}}{((\exists y')(\psi[y'/y]))}$ & $\alpha$-equivalence\\[2pt]
     3 & $\sequent{\psi}{\vec{x},y}{((\exists y')\psi)}$ & $\exists$ 2\\[2pt]
     4 & $\sequent{(\varphi\wedge\psi)}{\vec{x},y}{\psi}$ & $\wedge E1$\\[2pt]
     5 & $\sequent{(\varphi\wedge\psi)}{\vec{x},y}{((\exists y')\psi)}$ 
       & Cut 4, 3\\[2pt]
     6 & $\sequent{(\varphi\wedge\psi)}{\vec{x},y}{\varphi}$ & $\wedge E0$\\[2pt]
     7 & $\sequent{((\exists y)(\varphi \wedge \psi))}{\vec{x}}{\varphi}$
       & $\exists$ 6\\[2pt]
     8 & $\sequent{((\exists y)(\varphi \wedge \psi))}{\vec{x}}{%
            ((\exists y')\psi)}$
       & $\exists$ 5\\[2pt]
     9 & $\sequent{((\exists y)(\varphi\wedge\psi))}{\vec{x}}{%
       ((\varphi\wedge(\exists y')\psi))}$ & $\wedge$ 7, 8 \\[2pt]
     10 & $\sequent{((\exists y)(\varphi\wedge\psi))}{\vec{x}}{%
       ((\varphi\wedge(\exists y)(\psi[y/y']))}$ & $\alpha$-equivalence\\[2pt]
     11 & $\sequent{((\exists y)(\varphi\wedge\psi))}{\vec{x}}{%
       ((\varphi\wedge(\exists y)(\psi)}$ & rewrite 10
    \end{tabular}

\vfill
\eject

\subsubsection{Two Rules with $\wedge$ and $\vee$}\label{usefulAndOr}

\begin{theorem}\label{thm:tworuleand}
In Horn logic we can derive
$\displaystyle\frac{\sequent{\varphi}{\vec{x}}{\psi}}{\sequent{(\varphi\wedge \chi)}{\vec{x}}{(\psi\wedge \chi)}}.$
\end{theorem}

    \begin{tabular}{lll}
    1 & $\sequent{\varphi}{\vec{x}}{\psi}$ & Hypothesis\\[2pt]
    2 & $\sequent{(\varphi\wedge \chi)}{\vec{x}}{\varphi}$ & $\wedge$E0\\[2pt]
    3 & $\sequent{(\varphi\wedge \chi)}{\vec{x}}{\psi}$ & Cut 2, 1\\[2pt]
    4 & $\sequent{(\varphi\wedge \chi)}{\vec{x}}{\chi}$ & $\wedge$E1\\[2pt]
    5 & $\sequent{(\varphi\wedge \chi)}{\vec{x}}{(\psi\wedge \chi)}$ & $\wedge$I 3, 4\\
    \end{tabular}

\begin{theorem}\label{thm:tworuleor}
In coherent logic we can derive
$\displaystyle\frac{\sequent{\varphi}{\vec{x}}{\psi}}{\sequent{(\chi\vee \varphi)}{\vec{x}}{(\chi\vee \psi)}}.$
\end{theorem}

   \begin{tabular}{lll}
     1 & $\sequent{\varphi}{\vec{x}}{\psi}$ & Hypothesis\\
     2 & $\sequent{\psi}{\vec{x}}{(\chi\vee \psi)}$ & $\vee$I 1\\
     3 & $\sequent{\varphi}{\vec{x}}{(\chi\vee \psi)}$ & Cut 1, 2\\
     4 & $\sequent{\chi}{\vec{x}}{(\varphi\vee \psi)}$ & $\vee$I 0\\
     5 & $\sequent{(\chi\vee \varphi)}{\vec{x}}{(\chi\vee \psi)}$ & $\vee$4, 3
   \end{tabular}

\subsubsection{Distributive Rules}\label{distributive}

\begin{theorem}\label{thm:distrule}
In intuitionistic logic, we can derive the Distributive Rule \\ 
$\displaystyle\sequent{(\varphi\wedge (\psi\vee \chi))}{\vec{x}}{((\varphi\wedge \psi)\vee (\varphi\wedge \chi))}.$
\end{theorem}

   \begin{tabular}{lll}
   1 & $\sequent{(\varphi\wedge \psi)}{\vec{x}}{\left((\varphi\wedge \psi)\vee (\varphi\wedge \chi)\right)}$ & $\vee I0$\\[2pt]
   2 & $\sequent{\psi}{\vec{x}}{\left(\varphi\Rightarrow \left((\varphi\wedge \psi)\vee (\varphi\wedge \chi)\right)\right)}$ & $\Rightarrow$ 1\\[2pt]
   3 & $\sequent{(\varphi\wedge \chi)}{\vec{x}}{\left((\varphi\wedge \psi)\vee (\varphi\wedge \chi)\right)}$ & $\vee I1$\\[2pt]
   4 & $\sequent{\chi}{\vec{x}}{\left(\varphi\Rightarrow \left((\varphi\wedge \psi)\vee (\varphi\wedge \chi)\right)\right)}$ & $\Rightarrow$ 3\\[2pt]
   5 & $\sequent{(\psi\vee \chi)}{\vec{x}}{\left(\varphi\Rightarrow \left((\varphi\wedge \psi)\vee (\varphi\wedge \chi)\right)\right)}$ & $\wedge$ 2, 4\\[2pt]
   6 & $\sequent{(\varphi\wedge (\psi\vee \chi))}{\vec{x}}{\left((\varphi\wedge \psi)\vee (\varphi\wedge \chi)\right)}$ & $\Rightarrow$ 5\\[2pt]
  \end{tabular}


\begin{theorem}\label{thm:distruleconverse}
In coherent logic we can derive the converse of the Distributive Rule \\ 
$\displaystyle\sequent{((\varphi\wedge \psi)\vee (\varphi\wedge \chi))}{\vec{x}}{(\varphi\wedge (\psi\vee \chi))}$. 
\end{theorem}

   \begin{tabular}{lll}
   1 & $\sequent{(\varphi\wedge \psi)}{\vec{x}}{\varphi}$ & $\wedge E0$\\[2pt]
   2 & $\sequent{(\varphi\wedge \psi)}{\vec{x}}{\psi}$ & $\wedge E1$\\[2pt]
   3 & $\sequent{\psi}{\vec{x}}{(\psi\vee \chi)}$   & $\vee I0$\\[2pt]
   4 & $\sequent{(\varphi\wedge \psi)}{\vec{x}}{(\psi\vee \chi)}$ & Cut 2, 3\\[2pt]
   5 & $\sequent{(\varphi\wedge \psi)}{\vec{x}}{(\varphi\wedge (\psi\vee \chi))}$ & $\wedge$ 1, 4\\[5pt]
   6 & $\sequent{(\varphi\wedge \chi)}{\vec{x}}{\varphi}$ & $\wedge E0$\\[2pt]
   7 & $\sequent{(\varphi\wedge \chi)}{\vec{x}}{\chi}$ & $\wedge E1$\\[2pt]
   8 & $\sequent{\chi}{\vec{x}}{(\psi\vee \chi)}$   & $\vee I1$\\[2pt]
   9 & $\sequent{(\varphi\wedge \chi)}{\vec{x}}{(\psi\vee \chi)}$ & Cut 7, 8\\[2pt]
   10 & $\sequent{(\varphi\wedge \chi)}{\vec{x}}{(\varphi\wedge (\psi\vee \chi))}$ & $\wedge$ 6,9\\[5pt]
   11 & $\sequent{((\varphi\wedge \psi)\vee (\varphi\wedge \chi))}{\vec{x}}{(\varphi\wedge (\psi\vee \chi))}$ & $\vee$ 5, 10
   \end{tabular}

\vfill
\eject

\begin{theorem}
In coherent logic we can derive 
$\displaystyle\sequent{(\varphi\vee (\psi\wedge \chi))}{\vec{x}}{((\varphi\vee \psi)\wedge (\varphi\vee \chi)))}$. 
\end{theorem}

   \begin{tabular}{lll}
   1 & $\sequent{\varphi}{\vec{x}}{(\varphi\vee \psi)}$ & $\vee I0$\\[2pt]
   2 & $\sequent{\varphi}{\vec{x}}{(\varphi\vee \chi)}$ & $\vee I0$\\[2pt]
   3 & $\sequent{\varphi}{\vec{x}}{((\varphi\vee \psi)\wedge (\varphi\vee \chi))}$ & $\wedge$ 1, 2\\[7pt]
   4 & $\sequent{(\psi\wedge \chi)}{\vec{x}}{\psi}$ & $\wedge E0$\\[2pt]
   5 & $\sequent{\psi}{\vec{x}}{(\varphi\vee \psi)}$ & $\vee I1$\\[2pt]
   6 & $\sequent{(\psi\vee \chi)}{\vec{x}}{(\varphi\vee \psi)}$ & Cut 4, 5\\[2pt]
   7 & $\sequent{(\psi\wedge \chi)}{\vec{x}}{\chi}$ & $\wedge E1$\\[2pt]
   8 & $\sequent{\chi}{\vec{x}}{(\varphi\vee \chi)}$ & $\vee I1$\\[2pt]
   9 & $\sequent{(\psi\vee \chi)}{\vec{x}}{(\varphi\vee \chi)}$ & Cut 7, 8\\[2pt]
  10 & $\sequent{(\psi\wedge \chi)}{\vec{x}}{((\varphi\vee \psi)\wedge (\varphi\vee \chi))}$ & $\wedge$ 6, 9\\[7pt]
  11 & $\sequent{(\varphi\vee (\psi\wedge \chi))}{\vec{x}}{((\varphi\vee \psi)\wedge (\varphi\vee \chi))}$ & $\vee$ 3, 10\\
   \end{tabular}

\begin{theorem}
In intuitionistic logic we can derive
$\displaystyle\sequent{((\varphi\vee \psi)\wedge (\varphi\vee \chi))}{\vec{x}}{(\varphi\vee (\psi\wedge \chi))}$. 
\end{theorem}

   \begin{tabular}{lll}
   1 & $\sequent{(\varphi\wedge (\varphi\vee \chi))}{\vec{x}}{\varphi}$ & $\wedge E0$\\
   2 & $\sequent{\varphi}{\vec{x}}{(\varphi\vee (\psi\wedge \chi))}$ & $\vee I0$\\
   3 & $\sequent{(\varphi\wedge (\varphi\vee \chi))}{\vec{x}}{(\varphi\vee (\psi\wedge \chi))}$ & Cut 1,2 \\
   4 & $\sequent{\varphi}{\vec{x}}{((\varphi\vee \chi)\Rightarrow(\varphi\vee (\psi\wedge \chi)))}$ & $\Rightarrow$ 3\\[5pt]
   5 & $\sequent{(\psi\wedge \chi)}{\vec{x}}{(\varphi\vee (\psi\wedge \chi))}$ & $\vee I1$\\
   6 & $\sequent{\chi}{\vec{x}}{(\psi\Rightarrow(\varphi\vee (\psi\wedge \chi)))}$ & $\Rightarrow$ 5\\[5pt]
   7 & $\sequent{(\varphi\wedge \psi)}{\vec{x}}{\varphi}$ & $\wedge E0$\\
   8 & $\sequent{(\varphi\wedge \psi)}{\vec{x}}{(\varphi\vee (\psi\wedge \chi))}$ & Cut 7,2\\
   9 & $\sequent{\varphi}{\vec{x}}{(\psi\Rightarrow (\varphi\vee (\psi\wedge \chi)))}$ & $\Rightarrow$ 8\\[5pt]
   10 & $\sequent{(\varphi\vee \chi)}{\vec{x}}{(\psi\Rightarrow (\varphi\vee (\psi\wedge \chi)))}$ & $\vee$ 6, 9\\
   11 & $\sequent{(\psi\wedge (\varphi\vee \chi))}{\vec{x}}{(\varphi\vee (\psi\wedge \chi))}$ & $\Rightarrow$ 10\\
   12 & $\sequent{\psi}{\vec{x}}{((\varphi\vee \chi)\Rightarrow (\varphi\vee (\psi\wedge \chi)))}$ & $\Rightarrow$ 11\\
   13 & $\sequent{(\varphi\vee \psi)}{\vec{x}}{((\varphi\vee \chi)\Rightarrow (\varphi\vee (\psi\wedge \chi)))}$ & $\vee$ 4, 12\\
   14 & $\sequent{((\varphi\vee \psi)\wedge (\varphi\vee \chi))}{\vec{x}}{(\varphi\vee (\psi\wedge \chi))}$ & $\Rightarrow$ 13
   \end{tabular}

\subsubsection{Rules with $\vee$ and $\exists$}

\begin{theorem}
In coherent logic 
we can derive $\sequent{((\exists y)(\varphi\!\vee\!\psi))\!\!}{\vec{x}}{%
       ((\exists y)\varphi\vee(\exists y)\psi))}$.
\end{theorem}

    \begin{tabular}{lll}
    1 & $\sequent{(\exists y)\varphi}{\vec{x}}{(\exists y)\varphi}$ & ID\\[1pt]
    2 & $\sequent{\varphi}{\vec{x}, y}{(\exists y)\varphi}$ & $\exists$ 1\\[1pt]
    3 & $\sequent{\varphi}{\vec{x}, y}{%
           ((\exists y)\varphi\vee (\exists y)\psi)}$ & $\vee I0$ and Cut\\[1pt]
    4 & $\sequent{(\exists y)\psi}{\vec{x}}{(\exists y)\psi}$ & ID\\[1pt]
    5 & $\sequent{\psi}{\vec{x}, y}{(\exists y)\psi}$ & $\exists$ 4\\[1pt]
    6 & $\sequent{\psi}{\vec{x}, y}{%
           ((\exists y)\varphi\vee (\exists y)\psi)}$ & $\vee I1$ and Cut\\[1pt]
    7 & $\sequent{(\varphi\vee\psi)}{\vec{x}, y}{%
           ((\exists y)\varphi\vee (\exists y)\psi)}$ & $\vee$ 3, 6\\[1pt]
    8 & $\sequent{(\exists y)(\varphi\vee\psi)}{\vec{x}}{%
           ((\exists y)\varphi\vee (\exists y)\psi)}$ & $\exists$ 7\\[3pt]
    \end{tabular}

\vfill
\eject
\begin{theorem}
In coherent logic  we can derive
$\sequent{((\exists y)\varphi\vee(\exists y)\psi))}{\vec{x}}{%
       ((\exists y)(\varphi\vee\psi))}$.
\end{theorem}

    \begin{tabular}{lll}
    1 & $\sequent{(\exists y)(\varphi\vee\psi)}{\vec{x}}{%
           (\exists y)(\varphi\vee\psi)}$ & ID\\[1pt]
    2 & $\sequent{(\varphi\vee\psi)}{\vec{x}, y}{%
           (\exists y)(\varphi\vee\psi)}$ & $\exists$ 1\\[1pt]
    3 & $\sequent{\varphi}{\vec{x}, y}{\varphi\vee \psi}$ & $\vee$I0\\[1pt]
    4 & $\sequent{\varphi}{\vec{x}, y}{(\exists y)(\varphi\vee\psi)}$ 
           & Cut 3, 2\\[1pt]
    5 & $\sequent{(\exists y)\varphi}{\vec{x}}{(\exists y)(\varphi\vee\psi)}$ 
           & $\exists$ 4\\[1pt]
    6 & $\sequent{\psi}{\vec{x}, y}{\varphi\vee \psi}$ & $\vee$I1\\[1pt]
    7 & $\sequent{\psi}{\vec{x}, y}{(\exists y)(\varphi\vee\psi)}$ 
           & Cut 6, 2\\[1pt]
    8 & $\sequent{(\exists y)\psi}{\vec{x}}{(\exists y)(\varphi\vee\psi)}$ 
           & $\exists$ 7\\[1pt]
    9 & $\sequent{((\exists y)\varphi\vee(\exists y)\psi))}{\vec{x}}{%
       ((\exists y)(\varphi\vee\psi))}$ & $\vee$ 5, 8
    \end{tabular}

\subsubsection{Implication Proofs}\label{implication}

\begin{theorem}\label{thm:imp1}
In intuitionistic logic we can derive
   $\sequent{(\varphi\wedge (\varphi\Rightarrow\psi))}{\vec{x}}{\psi}$.
\end{theorem}

  \begin{tabular}{lll}
  1 & $\sequent{(\varphi\Rightarrow\psi)}{\vec{x}}{(\varphi\Rightarrow\psi)}$& ID\\
  2 & $\sequent{(\varphi\wedge(\varphi\Rightarrow\psi))}{\vec{x}}{\psi}$ & $\Rightarrow$ 1
  \end{tabular}
\smallskip

\begin{theorem}\label{thm:imp2}
In intuitionistic logic we can derive
   $\sequent{\varphi}{\vec{x}}{(\psi\Rightarrow (\varphi\wedge\psi))}$.
\end{theorem}

  \begin{tabular}{lll}
  1 & $\sequent{(\varphi\wedge\psi)}{\vec{x}}{(\varphi\wedge\psi)}$ & ID\\
  2 & $\sequent{\varphi}{\vec{x}}{(\psi\Rightarrow (\varphi\wedge\psi))}$ & $\Rightarrow$ 1
  \end{tabular}\vspace{5pt}
\smallskip

\begin{theorem}\label{thm:imp3}
In intuitionistic logic we can derive
   $\sequent{(\neg\varphi\vee \psi)}{\vec{x}}{(\varphi\Rightarrow \psi)}$.
\end{theorem}

  \begin{tabular}{lll}
  1 & $\sequent{(\neg\varphi\wedge\varphi}{\vec{x}}{\bot}$ & Theorem [\ref{thm:neg2}]\\
  2 & $\sequent{\bot}{\vec{x}}{\psi}$ & $\bot$\\
  3 & $\sequent{(\neg\varphi\wedge\varphi}{\vec{x}}{\psi}$ & Cut 1, 2\\
  4 & $\sequent{\neg\varphi}{\vec{x}}{(\varphi\Rightarrow\psi)}$ & $\Rightarrow$ 3\\
  5 & $\sequent{(\psi\wedge\varphi)}{\vec{x}}{\psi}$ & $\wedge E0$\\
  6 & $\sequent{\psi}{\vec{x}}{(\varphi\Rightarrow\psi)}$ & $\Rightarrow$ 5\\
  7 & $\sequent{(\neg\varphi\vee\psi)}{\vec{x}}{(\varphi\Rightarrow\psi)}$ & $\vee$ 4, 6
  \end{tabular}
\smallskip

\begin{theorem}
In classical logic we can derive
   $\sequent{(\varphi\Rightarrow\psi)}{\vec{x}}{(\neg\varphi\vee\psi)}$.
\end{theorem}

  \begin{tabular}{lll}
   1 &  $\sequent{\varphi\wedge(\varphi\Rightarrow \psi))}{\vec{x}}{\psi}$ & Theorem [\ref{thm:imp1}]\\
   2 &  $\sequent{\psi}{\vec{x}}{(\neg\varphi\vee \psi)}$ & $\vee I1$\\
   3 &  $\sequent{(\varphi\wedge(\varphi\Rightarrow\psi))}{\vec{x}}{(\neg\varphi\vee \psi)}$ & Cut 1, 2\\
   4 &  $\sequent{\varphi}{\vec{x}}{((\varphi\Rightarrow\psi)\Rightarrow (\neg\varphi\vee\psi))}$ 
     & $\Rightarrow$ 3\\[3pt]
   5 &  $\sequent{(\neg\varphi\wedge (\varphi\Rightarrow\psi))}{\vec{x}}{\neg\varphi}$ & $\wedge E0$\\
   6 &  $\sequent{\neg\varphi}{\vec{x}}{(\neg\varphi\vee \psi)}$ & $\vee I0$\\
   7 &  $\sequent{(\neg\varphi\wedge (\varphi\Rightarrow\psi))}{\vec{x}}{(\neg\varphi\vee \psi)}$ & Cut 5, 6\\
   8 &  $\sequent{\neg\varphi}{\vec{x}}{((\varphi\Rightarrow\psi)\Rightarrow (\neg\varphi\vee \psi))}$ 
           & $\Rightarrow$ 7\\
   9 &  $\sequent{(\varphi\vee\neg\varphi)}{\vec{x}}{((\varphi\Rightarrow\psi)\Rightarrow (\neg\varphi\vee \psi))}$
           & $\vee$ 4, 8\\[3pt]
  10 &  $\sequent{\top}{\vec{x}}{(\varphi\vee\neg\varphi)}$ & EM\\
  11 &  $\sequent{\top}{\vec{x}}{((\varphi\Rightarrow\psi)\Rightarrow (\neg\varphi\vee \psi))}$ & Cut 10, 9\\
  12 &  $\sequent{(\top\wedge (\varphi\Rightarrow\psi)}{\vec{x}}{(\neg\varphi\vee \psi)}$ & $\Rightarrow$ 11\\
  13 &  $\sequent{(\varphi\Rightarrow\psi)}{\vec{x}}{(\top\wedge (\varphi\Rightarrow\psi))}$ 
           & Theorem~[\ref{thm:topwedge}]\\
  14 &  $\sequent{(\varphi\Rightarrow\psi)}{\vec{x}}{(\neg\varphi\vee\psi)}$ & Cut 13, 12\\[3pt]
  \end{tabular}

Corollary: In classical logic $\sequentS{(\varphi\Rightarrow\psi)}{\vec{x}}{(\neg\varphi\vee\psi)}$.

\subsubsection{Negation Rules}\label{negationrules}

\begin{theorem}\label{thm:neg1}
In intuitionistic logic we can derive:
$\displaystyle 
  \frac{\underline{\sequent{(\varphi\wedge \psi)}{\vec{x}}{\bot}}}{\sequent{\psi}{\vec{x}}{\neg \varphi}}$
\end{theorem}

  \begin{tabular}{lll}
 Proof from top to bottom:\span\\
 1 & $\sequent{(\varphi\wedge \psi)}{\vec{x}}{\bot}$ & Hypothesis\\
 2 & $\sequent{\psi}{\vec{x}}{(\varphi\Rightarrow \bot)}$ & $\Rightarrow$ 1\\
 3 & $\sequent{\psi}{\vec{x}}{(\neg \varphi)}$ & Definition of $\neg$\\
Proof from bottom to top is similar.\span
  \end{tabular}

\begin{theorem}\label{thm:neg2}
In intuitionistic logic we can derive
$\displaystyle\sequent{(\varphi\wedge \neg \varphi)}{\vec{x}}{\bot}.$
\end{theorem}

  \begin{tabular}{lll}
    1 & $\sequent{\neg \varphi}{\vec{x}}{(\varphi\Rightarrow \bot)}$ & Definition of $\neg$\\
    2 & $\sequent{(\varphi\wedge \neg \varphi)}{\vec{x}}{\bot}$     & $\Rightarrow$ 1
  \end{tabular}\\[5pt]

\subsubsection{A Contradiction Rule}

\begin{theorem}
In intuitionistic logic we can derive:
 $\displaystyle\frac{\underline{\sequent{\top}{\vec{x}}{\neg \varphi}}}{\sequent{\varphi}{\vec{x}}{\bot}}.$
\end{theorem}

  \begin{tabular}{lll}
Proof from top to bottom:\span\\
    1 & $\sequent{\top}{\vec{x}}{\neg \varphi}$ & Hypothesis\\
    2 & $\sequent{\neg \varphi}{\vec{x}}{(\varphi\Rightarrow \bot)}$ & Definition of $\neg$\\
    3 & $\sequent{\top}{\vec{x}}{(\varphi\Rightarrow \bot)}$   & Cut 1, 2\\
    4 & $\sequent{(T\wedge \varphi)}{\vec{x}}{\bot}$ & $\Rightarrow$ 3\\
    5 & $\sequent{\varphi}{\vec{x}}{\varphi}$ & ID\\
    6 & $\sequent{\varphi}{\vec{x}}{\top}$ & $\top$\\
    7 & $\sequent{\varphi}{\vec{x}}{(\top\wedge \varphi)}$ & $\wedge$I 6, 5\\
    8 & $\sequent{\varphi}{\vec{x}}{\bot}$ & Cut 7, 4\\[5pt]
  Proof from bottom to top:\span\\
    1 & $\sequent{\varphi}{\vec{x}}{\bot}$ & Hypothesis\\
    2 & $\sequent{(\top\wedge \varphi)}{\vec{x}}{\varphi}$ & $\wedge$E1\\
    3 & $\sequent{(\top\wedge \varphi)}{\vec{x}}{\bot}$ & Cut 2, 1\\
    4 & $\sequent{\top}{\vec{x}}{(\varphi\Rightarrow \bot)}$ & $\Rightarrow$ 3\\
    5 & $\sequent{\top}{\vec{x}}{\neg \varphi}$ & Definition of $\neg$\vspace{3pt}
  \end{tabular}

\subsubsection{Double Negation Rules}\label{doublenegation}

\begin{theorem}\label{thm:negneg1}
In intuitionistic logic we can derive: 
$\sequent{\varphi}{\vec{x}}{\neg\neg\varphi}$
\end{theorem}

  \begin{tabular}{lll}
  1 & $\sequent{(\varphi\wedge \neg\varphi)}{\vec{x}}{\bot}$ & Theorem [\ref{thm:neg2}]\\
  2 & $\sequent{\varphi}{\vec{x}}{(\neg\varphi\Rightarrow \bot)}$ & $\Rightarrow$ 1\\
  3 & $\sequent{\varphi}{\vec{x}}{\neg\neg\varphi}$ & definition of $\neg$
  \end{tabular}\vspace{3pt}

\vfill
\eject

\begin{theorem}\label{thm:negneg2}
In classical logic we can derive:
      $\varphi$, $\sequent{\neg\neg\varphi}{\vec{x}}{\varphi}$.
\end{theorem}

  \begin{tabular}{lll}
  1 & $\sequent{(\neg\varphi\wedge \neg\neg\varphi)}{\vec{x}}{\bot}$ & Theorem [\ref{thm:neg2}]\\
  2 & $\sequent{\bot}{\vec{x}}{\varphi}$ & $\bot$\\
  3 & $\sequent{(\neg\varphi\wedge\neg\neg\varphi)}{\vec{x}}{\varphi}$ & Cut 1, 2\\
  4 & $\sequent{\neg\varphi}{\vec{x}}{(\neg\neg\varphi\Rightarrow \varphi)}$ & $\Rightarrow$ 3\\
  5 & $\sequent{(\varphi\wedge\neg\neg\varphi)}{\vec{x}}{\varphi}$ & $\wedge E0$\\
  6 & $\sequent{\varphi}{\vec{x}}{(\neg\neg\varphi\Rightarrow\varphi)}$ & $\Rightarrow$ 5\\
  7 & $\sequent{(\varphi\vee\neg\varphi)}{\vec{x}}{(\neg\neg\varphi\Rightarrow\varphi)}$ & $\vee$ 4, 6\\
  8 & $\sequent{\top}{\vec{x}}{(\varphi\vee\neg\varphi)}$ & EM\\
  9 & $\sequent{\top}{\vec{x}}{(\neg\neg\varphi\Rightarrow\varphi)}$ & Cut 8, 7\\
  10 & $\sequent{(\top\vee\neg\neg\varphi)}{\vec{x}}{\varphi}$ & $\Rightarrow$ 9\\
  11 & $\sequent{\neg\neg\varphi}{\vec{x}}{(\top\wedge \neg\neg\varphi)}$ & Theorem [\ref{thm:topwedge}]\\
  12 & $\sequent{\neg\neg\varphi}{\vec{x}}{\varphi}$ & Cut 11, 10
  \end{tabular}

\subsubsection{Contrapositive Rules}

\begin{theorem}\label{thm:contra1}
In intuitionistic logic we can derive: 
   $\sequent{(\varphi\Rightarrow\psi)}{\vec{x}}{(\neg\psi\Rightarrow\neg\varphi)}$.  
\end{theorem}

  \begin{tabular}{lll}
  1 & $\sequent{(\varphi\wedge (\varphi\Rightarrow\psi))}{\vec{x}}{\psi}$ & Theorem [\ref{thm:imp1}]\\
  2 & $\sequent{\psi}{\vec{x}}{\neg\neg\psi}$ & Theorem [\ref{thm:negneg1}]\\
  3 & $\sequent{(\varphi\wedge (\varphi\Rightarrow\psi))}{\vec{x}}{\neg\neg\psi}$ & Cut 1, 2\\
  4 & $\sequent{(\varphi\wedge (\varphi\Rightarrow \psi)\wedge \neg \psi)}{\vec{x}}{\bot}$ & 
            $\Rightarrow$ 3\\
  5 & $\sequent{((\varphi\Rightarrow\psi)\wedge \neg\psi)}{\vec{x}}{\neg\varphi}$ & $\Rightarrow$ 4\\
  6 & $\sequent{(\varphi\Rightarrow\psi)}{\vec{x}}{(\neg\psi\Rightarrow\neg\varphi)}$ & $\Rightarrow$ 5
  \end{tabular}

\begin{theorem}\label{thm:contra2}
In classical logic we can derive:
   $\sequent{(\neg\psi\Rightarrow\neg\varphi)}{\vec{x}}{(\varphi\Rightarrow\psi)}$.  
\end{theorem}

  \begin{tabular}{lll}
  1 & $\sequent{(\neg\psi\Rightarrow\varphi)}{\vec{x}}{(\neg\psi\Rightarrow\varphi)}$ & ID\\
  2 & $\sequent{((\neg\psi\Rightarrow\varphi)\wedge\neg\psi)}{\vec{x}}{\neg\varphi}$ & $\Rightarrow$ 1\\
  3 & $\sequent{((\neg\psi\Rightarrow\varphi)\wedge \neg\psi \wedge \varphi)}{\vec{x}}{\bot}$ & $\Rightarrow$ 2\\
  4 & $\sequent{((\neg\psi\Rightarrow\varphi)\wedge\varphi)}{\vec{x}}{\neg\neg\psi}$ & $\Rightarrow$ 3\\
  5 & $\sequent{\neg\neg \psi}{\vec{x}}{\psi}$ & Theorem [\ref{thm:negneg2}]\\
  6 & $\sequent{((\neg\psi\Rightarrow\varphi)\wedge\varphi)}{\vec{x}}{\psi}$ & $\Rightarrow$ 5\\
  7 & $\sequent{(\neg\psi\Rightarrow\varphi)}{\vec{x}}{(\varphi\Rightarrow \psi)}$ & $\Rightarrow$ 6\\
  \end{tabular}

\subsubsection{De Morgan's Rules}\label{demorgan}

\begin{theorem}\label{thm:demorganB}
In intuitionistic logic we can derive: 
   $\sequent{\neg(\varphi\vee \psi)}{\vec{x}}{(\neg \varphi\wedge \neg \psi)}$.  
\end{theorem}

  \begin{tabular}{lll}
  1 & $\sequent{\varphi}{\vec{x}}{(\varphi\vee \psi)}$ & $\vee I 0$\\
  2 & $\sequent{(\neg(\varphi\vee \psi)\wedge \varphi)}{\vec{x}}{(\neg(\varphi\vee \psi)\wedge (\varphi\vee \psi))}$ & Theorem [\ref{thm:tworuleor}]\\
  3 & $\sequent{(((\varphi\vee \psi)\Rightarrow \bot)\wedge (\varphi\vee \psi))}{\vec{x}}{\bot}$ 
    & Theorem [\ref{thm:imp1}]\\
  4 & $\sequent{(((\varphi\vee \psi)\Rightarrow \bot)\wedge \varphi)}{\vec{x}}{\bot}$         & Cut 2, 3\\
  5 & $\sequent{((\varphi\vee \psi)\Rightarrow \bot)}{\vec{x}}{\neg \varphi}$              & $\Rightarrow$ 4\\[5pt]
  6 & $\sequent{\psi}{\vec{x}}{(\varphi\vee \psi)}$   & $\vee I1$\\
  7 & $\sequent{(\neg(\varphi\vee \psi)\wedge \psi)}{\vec{x}}{(\neg(\varphi\vee \psi)\wedge (\varphi\vee \psi))}$ & Theorem [\ref{thm:tworuleand}]\\
  8 & $\sequent{(((\varphi\vee \psi)\Rightarrow \bot)\wedge \psi)}{\vec{x}}{\bot}$         & Cut 7, 3\\
  9 & $\sequent{((\varphi\vee \psi)\Rightarrow \bot)}{\vec{x}}{\neg \psi}$ & $\Rightarrow$ 8\\
  10 & $\sequent{\neg(\varphi\vee \psi)}{\vec{x}}{(\neg \varphi\wedge \neg \psi)}$ & $\wedge$ 5, 9
  \end{tabular}

\vfill
\eject

\begin{theorem}\label{thm:demorganA}
In intuitionistic logic we can derive: 
   $\sequent{(\neg\varphi\vee\neg\psi)}{\vec{x}}{\neg(\varphi\wedge\psi)}$.\vspace{-2pt}
\end{theorem}

  \begin{tabular}{lll}
   1 &  $\sequent{(\neg\varphi\wedge\varphi)}{\vec{x}}{\bot}$ & Theorem [\ref{thm:neg2}]\\[-.6pt]
   2 &  $\sequent{(\neg\varphi\wedge\varphi\wedge \psi)}{\vec{x}}{(\bot\wedge \psi)}$ & Theorem [\ref{thm:tworuleand}]\\[-.6pt]
   3 &  $\sequent{(\bot\wedge\psi)}{\vec{x}}{\bot}$ & $\wedge E0$\\[-.6pt]
   4 &  $\sequent{(\neg\varphi\wedge(\varphi\wedge\psi)}{\vec{x}}{\bot}$ & Cut 2, 3\\[-.6pt]
   5 & $\sequent{\neg\varphi}{\vec{x}}{((\varphi\wedge\psi)\Rightarrow \neg)}$ & $\Rightarrow$ 4\\[-.6pt]
   6 & $\sequent{\neg\varphi}{\vec{x}}{\neg(\varphi\wedge\psi)}$ & definition of $\neg$\\[-.6pt]
   7 & $\sequent{(\neg\psi\wedge\psi)}{\vec{x}}{\bot}$ & Theorem [\ref{thm:neg2}]\\[-.6pt]
   8 & $\sequent{(\neg\psi\wedge\psi\wedge\varphi)}{\vec{x}}{(\bot\wedge\varphi)}$ & Theorem [\ref{thm:tworuleand}]\\[-.6pt]
   9 & $\sequent{(\bot\wedge\varphi)}{\vec{x}}{\bot}$ & $\wedge E0$\\[-.6pt]
  10 & $\sequent{(\neg\psi\wedge\psi\wedge\varphi}{\vec{x}}{\bot}$ & Cut 8, 9\\[-.6pt]
  11 & $\sequent{\neg\psi}{\vec{x}}{((\varphi\wedge\psi)\Rightarrow\bot)}$ & $\Rightarrow$ 10\\[-.6pt]
  12 & $\sequent{\neg\psi}{\vec{x}}{\neg(\varphi\wedge\psi)}$ & definition of $\neg$\\[-.6pt]
  13 & $\sequent{(\neg\varphi\vee\neg\psi)}{\vec{x}}{\neg(\varphi\wedge\alpha)}$ & $\vee$ 6, 12\\[2pt]
  \end{tabular}

\begin{theorem}
In classical logic we can derive: 
   $\sequent{\neg(\varphi\wedge\psi)}{\vec{x}}{(\neg\varphi\vee\neg\psi)}$.\vspace{-2pt}
\end{theorem}

  \begin{tabular}{lll}
   1 & $\sequent{(\neg(\neg\varphi\vee\neg\psi))}{\vec{x}}{(\neg\neg\varphi\wedge\neg\neg\psi)}$ & 
         Theorem [\ref{thm:demorganB}]\\[-.6pt]
   2 & $\sequent{\neg\neg\varphi}{\vec{x}}{\varphi}$ & Theorem [\ref{thm:negneg2}] (uses EM)\\[-.6pt]
   3 & $\sequent{(\neg\neg\varphi\wedge\neg\neg\psi)}{\vec{x}}{(\varphi\wedge\neg\neg\psi)}$
           & Theorem [\ref{thm:tworuleand}]\\[-.6pt]
   4 & $\sequent{\neg\neg\psi}{\vec{x}}{\psi}$ & Theorem [\ref{thm:negneg2}] (uses EM)\\[-.6pt]
   5 & $\sequent{(\varphi\wedge\neg\neg\psi)}{\vec{x}}{(\varphi\wedge\psi)}$ & Theorem [\ref{thm:tworuleand}]\\[-.6pt]
   6 & $\sequent{(\neg\neg\varphi\wedge\neg\neg\psi)}{\vec{x}}{(\varphi\wedge\psi)}$ & Cut 3, 5\\[-.6pt]
   7 & $\sequent{(\neg(\neg\varphi\vee\neg\psi))}{\vec{x}}{(\varphi\wedge\psi)}$ & Cut 1, 7\\[-.6pt]
   8 & $\sequent{(\top\wedge (\neg(\neg\varphi\vee\neg\psi)))}{\vec{x}}{(\neg(\neg\varphi\vee\neg\psi))}$ 
          & Theorem [\ref{thm:topwedge}]\\[-.6pt]
   9 & $\sequent{(\top\wedge (\neg(\neg\varphi\vee\neg\psi)))}{\vec{x}}{(\varphi\wedge\psi)}$ & Cut 8, 7\\[-.6pt]
   10 & $\sequent{\top}{\vec{x}}{((\neg(\neg\varphi\vee\neg\psi)))\Rightarrow (\varphi\wedge\psi)}$ & $\Rightarrow$ 9\\[-.6pt]
   11 & $\sequent{(\neg(\neg\varphi\vee\neg\psi)))\Rightarrow (\varphi\wedge\psi))}{\vec{x}}{%
      (\neg(\varphi\wedge\psi)\Rightarrow\neg\neg(\neg\varphi\vee\neg\psi))}$ & 
          Theorem [\ref{thm:contra1}]\\[-.6pt]
   12 & $\sequent{\top}{\vec{x}}{(\neg(\varphi\wedge\psi)\Rightarrow\neg\neg(\neg\varphi\vee\neg\psi))}$ 
    & Cut 10, 11\\[-.6pt]
   13 & $\sequent{(\top\wedge\neg(\varphi\wedge\psi))}{\vec{x}}{(\neg\neg(\neg\varphi\vee\neg\psi))}$ 
        & $\Rightarrow$ 12\\[-.6pt]
   14  & $\sequent{(\neg(\varphi\wedge\psi))}{\vec{x}}{(\top\wedge (\neg(\varphi\wedge\psi)))}$
         & Theorem [\ref{thm:topwedge}]\\[-.6pt]
   15 & $\sequent{(\neg(\varphi\wedge\psi))}{\vec{x}}{(\neg\neg(\neg\varphi\vee\neg\psi))}$ 
          & Cut 13, 14\\[-.6pt]
   16 & $\sequent{(\neg\neg(\neg\varphi\vee\neg\psi))}{\vec{x}}{(\neg\varphi\vee\neg\psi)}$ 
       & Theorem [\ref{thm:negneg2}] (uses EM) \\[-.6pt]
   17 & $\sequent{(\neg(\varphi\wedge\psi))}{\vec{x}}{(\neg\varphi\vee\neg\psi)}$ & Cut 15, 16\\
  \end{tabular}

\begin{theorem}
In classical logic we can derive:
   $\sequent{\neg(\varphi\wedge\psi)}{\vec{x}}{(\neg\varphi\vee\neg\psi)}$.
\end{theorem}

  \begin{tabular}{lll}
  1 & $\sequent{\neg\neg\varphi}{\vec{x}}{\varphi}$ & Theorem [\ref{thm:negneg2}] (uses EM)\\[-0.6pt]
  2 & $\sequent{(\neg\neg\varphi\wedge \neg\neg\psi)}{\vec{x}}{(\varphi\wedge\neg\neg\psi)}$
          & Theorem [\ref{thm:tworuleand}]\\[-0.6pt]
  3 & $\sequent{\neg\neg\psi}{\vec{x}}{\psi}$ & Theorem [\ref{thm:negneg2}] (uses EM)\\[-0.6pt]
  4 & $\sequent{(\varphi\wedge\neg\neg\psi)}{\vec{x}}{(\varphi\wedge\psi)}$ &   
          Theorem [\ref{thm:tworuleand}]\\[-0.6pt]
  5 & $\sequent{(\neg\neg\varphi\wedge\neg\neg\psi)}{\vec{x}}{(\varphi\wedge\psi)}$ & Cut 2, 4\\[-0.6pt]
  6 & $\sequent{\neg(\neg\varphi\vee\neg\psi)}{\vec{x}}{(\neg\neg\varphi\wedge \neg\neg\psi)}$ 
       & Theorem [\ref{thm:demorganB}]\\[-0.6pt]
  7 & $\sequent{\neg(\neg\varphi\vee \neg\psi)}{\vec{x}}{(\varphi\wedge\psi)}$ & Cut 6, 5\\[-0.6pt]
  8 & $\sequent{\top}{\vec{x}}{((\neg(\neg\varphi\vee\neg\psi))\Rightarrow (\varphi\wedge \psi))}$
        & $\Rightarrow$ 7\\[-0.6pt]
  9 & $\sequent{(\neg(\neg\varphi\vee\neg\psi))\Rightarrow (\varphi\wedge \psi))}{\vec{x}}{%
       (\neg(\varphi\wedge\psi)\Rightarrow\neg\neg(\neg\varphi\vee\neg\psi))}$ 
       & Theorem [\ref{thm:contra1}]\\[-0.6pt]
  10 & $\sequent{\top}{\vec{x}}{%
     (\neg(\varphi\wedge\psi)\Rightarrow\neg\neg(\neg\varphi\vee\neg\psi))}$ & Cut 8, 9\\[-0.6pt]
  11 & $\sequent{(\top\wedge\neg(\varphi\wedge\psi))}{\vec{x}}{(\neg\neg(\neg\varphi\vee\neg\psi)))}$ 
      & $\Rightarrow$ 10\\[-0.6pt]
  12 & $\sequent{\neg(\varphi\wedge\psi)}{\vec{x}}{(\top\wedge \neg(\varphi\wedge\psi))}$ & 
         Theorem [\ref{thm:topwedge}]\\[-0.6pt]
  13 & $\sequent{\neg(\varphi\wedge\psi)}{\vec{x}}{\neg\neg(\neg\varphi\vee\neg\psi))}$ & Cut 12, 11\\[-0.6pt]
  14 & $\sequent{\neg\neg(\neg\varphi\vee\neg\psi))}{\vec{x}}{(\neg\varphi\vee\neg\psi)}$
       & Theorem [\ref{thm:negneg2}] (uses EM)\\[-0.6pt]
  15 & $\sequent{\neg(\varphi\wedge\psi)}{\vec{x}}{(\neg\varphi\vee\neg\psi)}$ & Cut 13, 14\\[-10pt]
  \end{tabular}

\vfill
\eject

\subsubsection{Resolution}

\begin{theorem} (\textbf{Intuitionistic Resolution Rule}):
In intuitionistic logic we can derive\vspace{-5pt}
\[\displaystyle\frac{\sequent{\varphi}{\vec{x}}{(\psi\vee \chi)}\hspace{20pt}\sequent{\varphi}{\vec{x}}{\neg \chi}}{\sequent{\varphi}{\vec{x}}{\psi}}\]
\end{theorem}

   \begin{tabular}{lll}
   1 & $\sequent{\varphi}{\vec{x}}{\psi\vee \chi}$ & Hypothesis\\[2pt]
   2 & $\sequent{\varphi}{\vec{x}}{\neg \chi}$  & Hypothesis\\[2pt]
   3 & $\sequent{\varphi}{\vec{x}}{(\psi\vee \chi)\wedge \neg \chi}$ & $\wedge I$ 1,2\\[2pt]
   4 & $\sequent{(\psi\vee \chi) \wedge (\neg \chi)}{\vec{x}}{(\psi\wedge \neg \chi)\vee (\chi\wedge \neg \chi)}$ & Theorem~[\ref{thm:distrule}]\\[2pt]
   5 & $\sequent{\varphi}{\vec{x}}{\big((\psi\wedge \neg \chi)\vee (\chi\wedge \neg \chi)\big)}$ & Cut 3, 4\\[2pt]
   6 & $\sequent{(\chi\wedge \neg \chi)}{\vec{x}}{\bot}$  & Theorem~[\ref{thm:neg2}]\\[2pt]
   7 & $\sequent{\big((\psi\wedge \neg \chi)\vee (\chi\wedge \neg \chi)\big)}{\vec{x}}{%
               \big((\psi\wedge \neg \chi)\vee \bot\big)}$ & 
      6, Theorem~[\ref{thm:tworuleor}]\\[2pt]
   8 & $\sequent{(\psi\wedge \neg \chi)\vee \bot}{\vec{x}}{(\psi\wedge \neg \chi)}$ 
            & Theorem~[\ref{thm:botvee}]\\[2pt]
   9 & $\sequent{(\psi\wedge \neg \chi)\vee (\chi\wedge \neg \chi)}{\vec{x}}{(\psi\wedge \neg \chi)}$ & Cut 7, 8\\[2pt]
  10 & $\sequent{\varphi}{\vec{x}}{(\psi\wedge \neg \chi)}$ & Cut 5, 9\\[2pt]
  11 & $\sequent{(\psi\wedge \neg \chi)}{\vec{x}}{\psi}$ & $\wedge E$\\[2pt]
  12 & $\sequent{\varphi}{\vec{x}}{\psi}$ & Cut 10, 11
   \end{tabular}

\begin{theorem}
 (\textbf{Coherent Resolution Rule}):
In coherent logic we can derive:
\[\displaystyle\frac{\sequent{\varphi}{\vec{x}}{(\psi\vee \chi)}\hspace{20pt}\sequent{(\varphi\wedge \chi)}{\vec{x}}{\bot}}{\sequent{\varphi}{\vec{x}}{\psi}}\]
\end{theorem}

   \begin{tabular}{lll}
   1 & $\sequent{(\varphi\wedge \psi)}{\vec{x}}{\psi}$    & $\wedge E$\\[2pt]
   2 & $\sequent{(\varphi\wedge \chi)}{\vec{x}}{\bot}$ & Hypothesis\\[2pt]
   3 & $\sequent{\bot}{\vec{x}}{\psi}$         & $\bot$\\[2pt]
   4 & $\sequent{(\varphi\wedge \chi)}{\vec{x}}{\psi}$    & Cut 2, 3\\[2pt]
   5 & $\sequent{((\varphi\wedge \psi)\vee ((\varphi\wedge \chi))}{\vec{x}}{\psi}$ & $\vee$ 1, 4\\[2pt]
   6 & $\sequent{(\varphi\wedge (\psi\vee \chi))}{\vec{x}}{(\varphi\wedge \psi)\vee (\varphi\wedge \chi)}$ 
         & Theorem [\ref{thm:distrule}]\\[2pt]
   7 & $\sequent{(\varphi\wedge (\psi\vee \chi))}{\vec{x}}{\psi}$ & Cut 5, 6\\[2pt]
   8 & $\sequent{\varphi}{\vec{x}}{\varphi}$ & Id\\[2pt]
   9 & $\sequent{\varphi}{\vec{x}}{\psi\vee \chi}$ & Hypothesis\\[2pt]
   10 & $\sequent{\varphi}{\vec{x}}{(\varphi\wedge (\psi\vee \chi))}$ & $\wedge$ 8, 9\\[2pt]
   11 & $\sequent{\varphi}{\vec{x}}{\psi}$ & Cut 10, 7
  \end{tabular}

\vfill
\eject

\subsection{Semantic Proofs}\label{sec:subprop}

\subsubsection{Substitution Properties}\label{substitution-properites}

Inductively define $\interp{\vec{x}.\,t}_M:M(X_1)\times\cdots\times M(X_m)\to M(B)$
by: (1)~if $t=x_j$ for  $1\leq j \leq m$, then $\interp{\vec{x}.\,t}_M=\pi_j$, the
product projection; (2)~if $t=f(t_1, \dots, t_n)$ with $t_i:A_i$, then
$\interp{\vec{x}.\,t}_M$ is the following composite 
where $\interp{\vec{x}.\,t_i}_M:M(X_1)\times\cdots\times M(X_m)\to M(A_i)$.\vspace{-12pt}
\[\xymatrix{
M(X_1)\times\cdots\times M(X_m)\ar[rrr]^{\left(\,\interp{\vec{x}.\,t_1}_M,\,\dots,\, \interp{\vec{x}.\,t_n}_M\right)}
         \ar[drrr]_(.55){\interp{\,\vec{x}.f(t_1,\dots,\, t_n)}_M\hphantom{AAAA}}
     &&&M(A_1)\times\cdots\times M(A_n)\ar[d]^(.5){M(f)}\\
&&& M(B).}\vspace{-8pt}\]

extend to formulae-in-context.

\begin{theorem}\label{substitution-property-terms}
\textit{Substitution Property for Terms}  (\cite{Elephant} Lemma D1.2.4): If $\vec{y}$ is a suitable
context for a term $t:B$,   $[\vec{s}/\vec{y}\,]$ is substitution, $y_i:Y_i$,
and $\vec{x}$ is a context suitable for $\vec{s}$, then 
$\interp{\vec{x}.\left(t[\vec{s}/\vec{y}\,]\right)}$ can be computed using composition:\vspace{-15pt}
\[\xymatrix{
M(X_1)\times\cdots\times M(X_m)\ar[rrr]^{\left(\,\interp{\vec{x}.\,s_1}_M,\,\dots,\, \interp{\vec{x}.\,s_n}_M\right)}
  \ar[drrr]_(.55){\interp{\,\vec{x}.\,\left(t[\vec{s}/\vec{y}\,]\right)}_M\hphantom{AAAAAAA}}
           &&& M(Y_1)\times\cdots\times M(Y_n)\ar[d]^{\interp{\,\vec{y}.\,t}_M}\\
&&& M(B)}\]
\end{theorem}

\myproof 
 If $t=y_i$ for $1\leq i\leq n$, then 
$\vec{x}.\big(t[\vec{s}/\vec{y}]\big)=\vec{x}.\big(y_i[\vec{s}/\vec{y}]\big)=\vec{x}.\,s_i$
and the  diagram commutes since 
$\interp{\vec{y}.\,t}_M=\interp{\vec{y}.y_i}_M=\pi_i$ and 
$\interp{\,\vec{x}.\,\left(y_i[\vec{s}/\vec{y}]\right)}_M =\interp{\,\vec{x}.\,s_i}_M$.
If $t=f(t_1, \dots, t_m)$, then 
$t[\vec{s}/\vec{y}\,] = f(t_1[\vec{s}/\vec{y}\,],\,\dots,\, t_m[\vec{s}/\vec{y}\,])$.  
The definition of substitution justifies the first equality of 
\[\interp{\vec{x}.t[\vec{s}/\vec{y}\,]} = \interp{\vec{x}.f(t_1[\vec{s}/\vec{y}\,],
    \,\dots,\, t_m[\vec{s}/\vec{y}\,])} = M(f)\circ \left(\interp{t_1[\vec{s}/\vec{y}\,]},
   \,\dots,\,\interp{t_n[\vec{s}/\vec{y}\,]}\right)\]
and the definition of semantics justifies the second.
By induction, 
$\interp{t_i[\vec{s}/\vec{y}\,]}=\interp{\vec{y}.t_i}\circ \left(\interp{\vec{x}.s_1},\,\dots,\,\interp{\vec{x}.s_n}\right)$.  Hence,
\[\interp{\vec{x}.t[\vec{s}/\vec{y}]} =
    M(f)\circ \left(\interp{\vec{y}.s_1},\,\dots,\,\interp{\vec{y}.s_m}\right)\circ
   \left(\interp{\vec{x}.s_1},\,\dots,\,\interp{\vec{x}.s_n}\right)
= \interp{\vec{y}.f(s_1, \dots, s_m)}\circ
     \left(\interp{\vec{x}.s_1},\,\dots,\,\interp{\vec{x}.s_n}\right)
\]
\hfill \enpr

\begin{theorem}
\textit{Substitution Property for Formulae} (\cite{Elephant} Lemma D1.2.7): If 
$\vec{y}$ is a suitable context for a formula $\varphi$,
$[\vec{s}/\vec{y}\,]$ is a substitution, and $\vec{x}$ is suitable for $\vec{s}$, 
then $\interp{\vec{x}.(\varphi[\vec{t}/\vec{y}\,])}$ can be computed using a pullback:
\[\xymatrix{
\interp{\vec{x}.(\varphi[\vec{s}/\vec{y}\,])}_M\ar@{->}[rrr]\ar@{|->}[d] 
      &&& \interp{\vec{y}.\varphi}_M\ar@{|->}[d]\\
M(X_1)\times\cdots\times M(X_m)\ar@{->}[rrr]_{(\interp{\vec{x}.s_1}_M,\,\dots,\,\interp{\vec{x}.s_n}_M)}
      &&& M(B_1)\times\cdots\times M(B_n)}\]
\end{theorem}
\vfill
\eject

\end{document}